\documentclass[10pt,letterpaper]{article}
\usepackage[top=0.85in,left=2.75in,footskip=0.75in]{geometry}

\usepackage{lmodern}
\usepackage{xcolor}
\usepackage{graphicx}
\usepackage{pstricks-add}
\usepackage{lscape}
\usepackage{float}
\usepackage{capt-of}
\usepackage{subcaption}
\usepackage{soul}
\usepackage{afterpage}
\usepackage{floatpag}
\usepackage{multirow}

\usepackage{adjustbox}
\usepackage{array}
\usepackage{csquotes}
\usepackage{tikz}
\usetikzlibrary{shapes.geometric, arrows.meta, positioning, shadows}
\usepackage{amsmath} 
\usepackage{algorithm} 
\usepackage{algpseudocode}
\usepackage{colortbl}
\usepackage{booktabs}
\usepackage{comment}
\usepackage{datetime}
\usepackage[normalem]{ulem}
\usepackage{enumitem}
\usepackage{changepage} 
\tikzstyle{startstop} = [rectangle, rounded corners, minimum width=2cm, minimum height=1cm, text centered, draw=black, fill=red!30, drop shadow]
\tikzstyle{io} = [trapezium, trapezium left angle=70, trapezium right angle=110, minimum width=3cm, minimum height=0.5cm, text centered, draw=black, fill=blue!30, drop shadow]
\tikzstyle{process} = [rectangle, minimum width=3cm, minimum height=1cm, text centered, draw=black, fill=orange!30, drop shadow]
\tikzstyle{decision} = [diamond, aspect=1, minimum width=1cm, minimum height=0.5cm, text centered, draw=black, fill=green!30, drop shadow]
\tikzstyle{arrow} = [thick,-Triangle,>=stealth]

 \usepackage{epsfig}
 \usepackage{geometry}
\usepackage{booktabs}
\usepackage{siunitx}
\usepackage{lipsum}
\usepackage{fancyhdr}

\usepackage{amssymb}
\usepackage{amsmath}
\usepackage{amsfonts}
\usepackage{hyperref}
 \hypersetup{
colorlinks=true,
linkcolor=blue,
filecolor=blue,
citecolor = blue,
urlcolor=blue,
}

\begin{document}

\vspace*{0.2in}

\begin{flushleft}
{\Large
\textbf\newline{Advanced Vision Transformers and Open-Set Learning for Robust Mosquito Classification: A Novel Approach to Entomological Studies}
}


Ahmed Akib Jawad Karim\textsuperscript{1},
Muhammad Zawad Mahmud\textsuperscript{1},
Riasat Khan\textsuperscript{1},

\bigskip
\textbf{1} Electrical and Computer Engineering, North South University, Dhaka, Bangladesh

\bigskip

\end{flushleft}


\section*{Abstract}
Mosquito-related diseases pose a significant threat to global public health, necessitating efficient and accurate mosquito classification for effective surveillance and control. This work presents an innovative approach to mosquito classification by leveraging state-of-the-art vision transformers and open-set learning techniques. A novel framework has been introduced that integrates Transformer-based deep learning models with comprehensive data augmentation and preprocessing methods, enabling robust and precise identification of ten mosquito species. The Swin Transformer model achieves the best performance for traditional closed-set learning with 99.60\% accuracy and 0.996 F1 score. The lightweight MobileViT technique attains an almost equivalent accuracy of 98.90\% with significantly reduced parameters and model complexities. Next, the applied deep learning models' adaptability and generalizability in a static environment have been enhanced by using new classes of data samples during the inference stage that have not been included in the training set. The proposed framework's ability to handle unseen classes like insects similar to mosquitoes, even humans, through open-set learning further enhances its practical applicability employing the OpenMax technique and Weibull distribution. The traditional CNN model, Xception, outperforms the latest transformer with higher accuracy and F1 score for open-set learning. The study's findings highlight the transformative potential of advanced deep-learning architectures in entomology, providing a strong groundwork for future research and development in mosquito surveillance and vector control. The implications of this work extend beyond mosquito classification, offering valuable insights for broader ecological and environmental monitoring applications.

\section*{Author Summary}
Mosquitoes are major vectors for diseases like malaria, dengue, and Zika, causing widespread health challenges globally. Identifying mosquito species accurately is critical for controlling these diseases. Traditionally, mosquito classification is done manually, which is labor-intensive and time-consuming. To address this, we have developed an automated system that uses advanced deep learning models, specifically vision transformers, to classify mosquitoes into ten distinct species, including Aedes, Anopheles, and Culex, using a dataset of 5000 images. Our approach not only improves mosquito identification in static environments but also introduces a novel feature called open-set learning, enabling the system to detect and flag insects or species that were not part of the training dataset. This makes the system highly adaptable for real-world applications where unexpected or new mosquito species may appear. By combining these cutting-edge technologies, our work aims to support more effective mosquito surveillance, aiding public health initiatives and disease prevention efforts, particularly in regions most affected by mosquito-borne illnesses. 

\section{Introduction}
\label{intro}

Vector-borne diseases are illnesses resulting from pathogens, including bacteria, viruses, or parasites, which are transmitted to humans through the bites of vectors, especially mosquitoes. Mosquitoes are tiny insects well-known for their capacity to spread illness to people and animals through bites \cite{10.3389/fitd.2023.1240420}. Mosquitoes are responsible for transmitting over 70$\%$ of vector-borne diseases globally \cite{WHO:2020}. There exist various debilitating diseases due to exposure to mosquito bites. For instance, Malaria is the deadliest disease transmitted by mosquitoes, accounting for more than 0.60 million deaths and 250 million cases worldwide in 2022. Dengue ranks second, with over 0.40 million deaths (mostly in Asia, particularly severe in Bangladesh) \cite{icddrb:2020}. It occurs due to the bite of Aedes alobopictus and Aedes aegypti species of mosquitoes~\cite{cdc:2023}.

Recent advances in artificial intelligence (AI) and effective computer vision techniques provide significant progress in mosquito surveillance and monitoring, which is essential in the battle against vector-borne illnesses. Conventional identification and classification of mosquitoes include manual counting, entomological and microscopic observations, PCR and DNA tests after trapping from insecticide-treated nets, bednets, etc. These traditional strategies are inefficient, time-consuming and expensive due to the involvement of complex thermal cycling machines, medical entomologists and public health experts. Due to diminutive bodies, rapid movement, complex body landmarks, and striking similarity in the appearance of mosquitoes with that of other flies, the detection and classification of mosquitoes is exceptionally challenging. AI with visual sensors and computing facilities can simplify the difficult process of identifying mosquito species from trapped samples. For instance, deep learning models such as convolutional neural networks have illustrated impressive precision in categorizing mosquitoes captured via traps or cameras \cite{minakshi2020automating}. Additionally, AI-driven systems can analyze various data streams, including weather patterns and historical disease occurrences, to predict mosquito population dynamics and the risks of disease transmission. It is essential to accurately estimate mosquito vector population density on a large scale and insignificant processing time for preventing infectious diseases and deploying targeted interventions.

In this work, an AI-based automatic mosquito classification system has been developed using a comprehensive dataset of ten species, advanced vision Transformer models and open-set learning techniques. The primary contributions of this work are as follows:

\begin{itemize}
    \item A significant contribution of this research is to develop a comprehensive mosquito dataset by merging four publicly available datasets. This consolidated dataset encompasses ten distinct classes, including Aedes aegypti, Aedes albopictus, Anopheles albimanus, Anopheles arabiensis, Anopheles atroparvus, Anopheles coluzzi, Anopheles farauti, Anopheles freeborni, Anopheles stephensi, and Culex quinquefasciatus with 1749 sample images, including partially damaged and smashed images of mosquitoes. 
    \item Advanced data preprocessing and augmentation techniques are performed to balance the dataset. 
   \item Transfer learning and Transformer-based deep learning models, e.g., Xception, ViT, CvT, MobileViT and Swin, have been applied to classify the mosquito species into ten categories in traditional closed-set static environments. 
    \item Additionally, open-set learning techniques based on the OpenMax layer approach and Weibull distribution are integrated into the applied models, enabling them to identify and categorize images outside the predefined mosquito classes as ``unknown." The core principle involves calculating class-specific confidence scores based on the distance between a sample's activation vector and the class mean activation vector, incorporating a Weibull distribution for robust uncertainty modeling. 
    \item The novelty of this work is to apply Transformer-based deep learning models and open-set learning techniques to classify mosquitoes using a comprehensive mosquito dataset of ten individual species.
\end{itemize}

\section{Related work}

The morphological characterization is important for accurate mosquito species classification. In recent times, significant work has been performed on AI-based automatic mosquito detection to aid taxonomic experts and public health professionals. For this reason, many researchers have shown their interest in classifying mosquitoes using deep learning approaches. Most of these used various pretrained classification and transformer models to classify mosquitoes through images and wingbeat sounds. Some recent articles on AI-based automatic mosquito detection are briefly discussed in the following paragraphs.

Rustam et al.~\cite{rustam2022vector} used machine learning algorithms and innovative RIFS feature selection approaches to classify vector mosquito images for disease epidemiology. The authors classified two disease-spreading classes of mosquitoes, Aedes and Culex, using an open-source dataset from the IEEE data port. VGG16 model with the RIFS feature selection approach accomplished 99.2$\%$ accuracy with 0.986 F1 coefficient. A mosquito categorization scheme using transfer learning was developed by Isawasan et al.~\cite{isawasan2023aprotocol}. The authors classified two species of Aedes mosquito with a public database of 1205 images. They used five CNN models with transfer learning approaches (ResNet50, Xception, InceptionV3, MobileNetV2, and VGG16), and all the applied models achieved accuracies of more than 90$\%$ with VGG16 attaining a maximum accuracy of 98.25$\%$.  

Siddiqui and Kayte~\cite{siddiqui2023transfer} applied various deep learning frameworks for mosquito classification. They collected a dataset of six species of mosquitoes from website search engines employing web-scrapping techniques. After augmentation, the dataset contained a total of 5400 images. The applied custom CNN and VGG16 models attained the maximum accuracies of 97.17$\%$ and 85.75$\%$, respectively. A comparative analysis of conventional and deep learning techniques for mosquito species categorization was investigated by Okayasu et al.~\cite{okayasu2019vision}. The authors employed 7200 images clicked through camera and three types of mosquito classes. The conventional SIFT method achieved the highest accuracy of 0.824. In contrast, the ResNet deep learning method had the highest of 0.955, demonstrating the effectiveness of deep classification in the categorization of mosquito species. 

Lee et al.~\cite{lee2023deep} investigated deep learning-integrated picture categorization for principal mosquito species in Korea. They used a dataset of 12 classes and eleven mosquito species comprising 7096 images after augmentation. The Swin transformer hybrid model offered the best performance with an F1 score of 0.971 and 96.5$\%$ precision. Zhao et al.~\cite{zhao2022swin} initiated mosquito species identification utilizing various deep learning techniques. The dataset consists of seven genera and 17 species, with a total of 9900 mosquito images after augmentation. The Swin transformer model accomplished the maximum accuracy, which is $96.3\%$. YOLOv5 and DETR gave good competition to Swin transformers with an accuracy of $94.3\%$ and $93.3\%$, respectively.

Omucheni et al.~\cite{omucheni2023rapid} used the Raman spectroscopy method and deep learning classification approaches to identify mosquito species. The authors used 397 mosquito images, including Anopheles gambiae and Anopheles arabiensis. The SVM model with a quadratic kernel performed best with 93$\%$ accuracy and 0.90 sensitivity. Rony et al.~\cite{rony2023mosquito} investigated the identification of mosquito species by combining CNN and SVM with the help of a wingbeat frequency audio dataset. The applied models were trained using a large variety of data because of the acoustic sensitivity and the reliance of mosquito wing flapping on the weather and surroundings. Finally, the combined SVM and CNN model acquired 99.46$\%$ accuracy and 0.99 F1 score. 

Yin et al.~\cite{yin2023deep} used a deep learning-based pipeline for classifying and detecting mosquitoes based on their wingbeat noises. The data was collected from Medical Entomology of the Tropical Medicine facility of a Thailand university, which contained the wingbeat sound of four species of mosquitoes. The combined CNN and LSTM models secured the best performance with 61$\%$ precision and a 0.62 F1 score. After applying multiple audio preprocessing techniques, the classification accuracy improved to 90$\%$. An ensemble of various convolutional neural networks for new species recognition in mosquito species identification was done by  Goodwin et al.~\cite{goodwin2021mosquito}. The authors used a dataset containing 16 classes. Among the models they used, the CNN model Xception attained the best result for both closed-set and open-set. For the closed-set, it gave an accuracy and F1 score of 0.974 and 0.966, respectively. The Xception model's accuracy and F1 score were reduced to 0.8906 and 0.7974 for the open-set novel class, respectively. 

Pise and Patil~\cite{pise2023deep} used a deep transfer learning framework for the multi-class classification of vector mosquito species of three classes. The dataset comprised 2,648 images, among which Aedes aegypti had 900 images, Anopheles stephensi had 540 images, and Culex Guinquefasciatus had 1208 images. GoogLeNet demonstrated superior performance, achieving a classification accuracy of 92.5\% with feature extraction and 96\% with fine-tuning of the corresponding hyperparameters.

The above studies demonstrate significant advancements in mosquito classification using various machine learning and deep learning approaches, with a particular emphasis on static environments. Several researchers have utilized traditional machine learning algorithms, convolutional neural networks (CNNs), and transformer models to classify mosquito species based on images and wingbeat sounds. In this constrained setup, training and testing of the models are performed with the same labels, leading to poor performance when the models encounter a new test class not present in the training samples. Due to mosquitoes' close morphological characterization similarity with other flies and insects, such as double wings, antennae, tri-paired legs, and compound-structured eyes, open-set learning-based classification should be investigated. In this flexible, non-stationary situation, unknown testing samples may be present, which were not declared during the training stages. This realistic setup is very suitable for deployment in real applications, enhancing robustness and usability. This work explores the potential of vision transformer models with open-set learning for improving the accuracy and adaptability of mosquito classification systems, contributing to better disease control and epidemiology efforts.

\section{Methodology}

The proposed vision transformers and open-set learning for the mosquito classification system are described in detail in this section. 

\subsection{Dataset}

\subsubsection{Formation of the Dataset}
\begin{figure}[htbp]
  \centering
  \begin{subfigure}{0.19\textwidth}
    \centering
    \includegraphics[width=\linewidth]{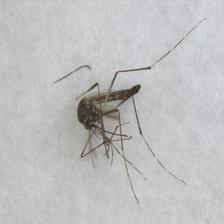}
    \caption{Ae. Aegypti}
  \end{subfigure}
  \begin{subfigure}{0.19\textwidth}
    \centering
    \includegraphics[width=\linewidth]{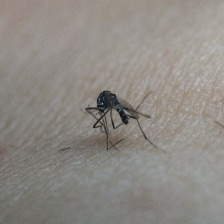}
    \caption{Ae. Albopictus}
  \end{subfigure}
  \begin{subfigure}{0.19\textwidth}
    \centering
    \includegraphics[width=\linewidth]{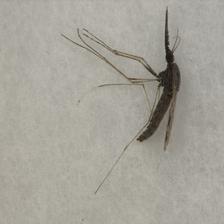}
    \caption{An. Albimanus}
  \end{subfigure}
  \begin{subfigure}{0.19\textwidth}
    \centering
    \includegraphics[width=\linewidth]{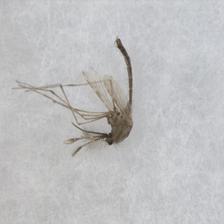}
    \caption{An. Arabiensis }
  \end{subfigure}
  \begin{subfigure}{0.19\textwidth}
    \centering
    \includegraphics[width=\linewidth]{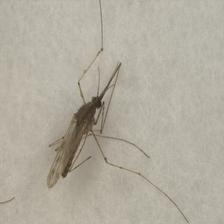}
    \caption{An. Atroparvus}
  \end{subfigure}
  
  \begin{subfigure}{0.19\textwidth}
    \centering
    \includegraphics[width=\linewidth]{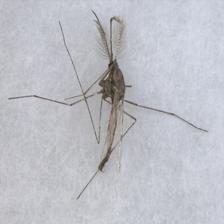}
    \caption{An. Coluzzi}
  \end{subfigure}
  \begin{subfigure}{0.19\textwidth}
    \centering
    \includegraphics[width=\linewidth]{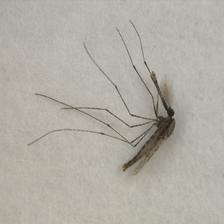}
    \caption{An. Farauti}
  \end{subfigure}
  \begin{subfigure}{0.19\textwidth}
    \centering
    \includegraphics[width=\linewidth]{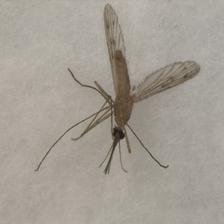}
    \caption{An. Freeborni}
  \end{subfigure}
  \begin{subfigure}{0.19\textwidth}
    \centering
    \includegraphics[width=\linewidth]{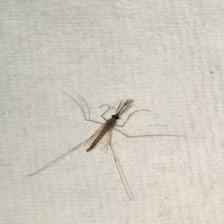}
    \caption{An. Stephensi}
  \end{subfigure}
  \begin{subfigure}{0.19\textwidth}
    \centering
    \includegraphics[width=\linewidth]{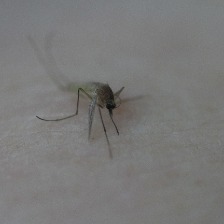}
    \caption{Cx. Quinquefasciatus}
  \end{subfigure}
  \caption{Sample images of various classes of mosquitoes of the comprehensive dataset used in this work.}
  \label{fig:dataset}
\end{figure}

The comprehensive dataset of this work has been constructed by combining various open-source image datasets utilized in related mosquito classification articles, e.g., image dataset of Aedes and Culex mosquito species~\cite{m05g-mq78-20}, mosquito-on-human-skin~\cite{ong2022data}, dataset of vector mosquito images~\cite{pise2022dataset} and malaria vector mosquito images~\cite{couret2020data}. The proposed dataset contains images of ten species of mosquitoes: two species of Aedes, seven of Anopheles and one Culex type. The images of the two classes, Aedes albopictus and Culex quinquefasciatus, contain both whole and smashed/crushed mosquitoes, which can be described as partial mosquitoes, resulting in only parts of the insect being visible or intact. Table~\ref{dataset_split_table} represents the exact numbers of images for various mosquito classes. A sample image of each class is shown in Figure~\ref{fig:dataset}. Aedes is represented as Ae, Anopheles as An, and Culex as Cx in the caption of the figure.

\begin{table}[ht] 
\caption{Number of images of various classes of mosquitoes of the employed dataset}
\label{dataset_split_table}
\begin{center}
\begin{tabular}{ccc}

\hline
Type       &       Total without Augmentation & Total with Augmentation\\
\hline  
Aedes aegypti       & 219     & 500   \\
Aedes albopictus      &  500   & 500    \\
Anopheles albimanus       & 64      & 500  \\
Anopheles arabiensis       & 193     & 500   \\
Anopheles atroparvus      & 36        & 500 \\
Anopheles coluzzi      & 74     & 500 \\
Anopheles farauti     & 57    & 500  \\
Anopheles freeborni     & 97    & 500 \\
Anopheles stephensi   &  9   & 500 \\
Culex quinquefasciatus      & 500  & 500 \\
\hline
Total      & 1749  & 5000 \\
\hline
\end{tabular}
\end{center}
\end{table}

\subsubsection{Dataset Preprocessing}
The data preprocessing approach was applied to expand the dataset by introducing slightly modified versions of existing images or generating new artificial data from the current images. This process acts as a regularizer, mitigating overfitting during model training~\cite{shorten2019survey}. Data augmentation techniques are strongly connected to oversampling in this context, allowing for an enhanced and balanced dataset. In this work, additional images were generated using standard augmentation techniques, including rotation (limited to 20 degrees), width and height shifts (0.20), shear (0.20), zoom (0.20), horizontal flips, and nearest fill mode via the \texttt{ImageDataGenerator} framework~\cite{franccois2015keras}.

\begin{figure}[h]
  \centering
  \includegraphics[width=0.9\linewidth]{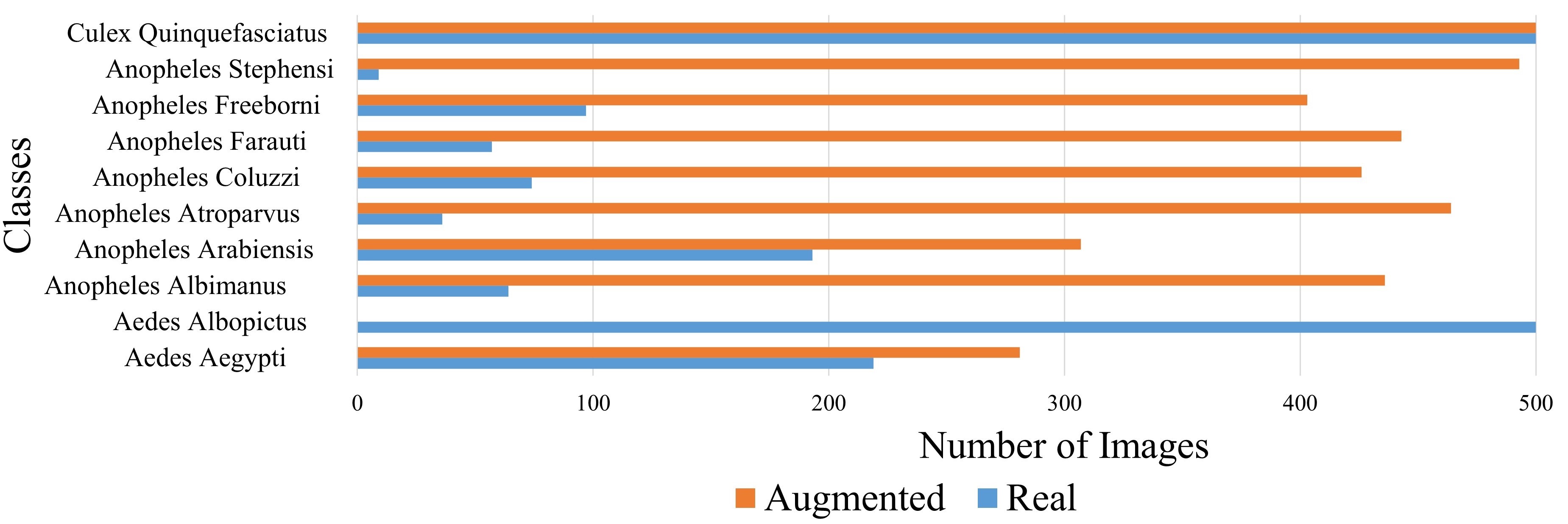}
  \caption{Number of augmented images of various mosquito classes in the dataset.}
  \label{fig:data_aug}
\end{figure}

The choice to limit the rotation to 20 degrees was intentional to preserve the natural wing positions and antennae orientation of mosquitoes, as these anatomical features are essential for accurate species classification. Larger rotations or deformations could distort these key features, negatively affecting the model's ability to distinguish between species. Additionally, the augmentation parameters such as shifts and zoom were set at 20\% to introduce sufficient variation without significantly altering the mosquito's structural integrity. This augmentation approach is widely considered a well-established standard in the field, balancing the need for diversity while maintaining the essential characteristics of the objects of interest. Figure~\ref{fig:data_aug} presents the number of augmented images for each mosquito class. 

\subsubsection{Dataset Splitting}

After data preprocessing and augmentation, the comprehensive dataset of 5000 images was split into training, validation, and test sets. Initially, the dataset was divided into a training set and a test set using an 80:20 ratio, resulting in 4000 images in the training set and 1000 images in the test set. The 4000 training images were then further split into a training set and a validation set, using another 80:20 ratio. This resulted in 3200 images for the final training set and 800 images for the validation set.

Thus, the final dataset consisted of 3200 images for training, 800 images for validation, and 1000 images for testing. This stratified splitting ensured that each mosquito species was proportionally represented across all sets, maintaining class balance throughout. During the training phase, the training set (3200 images) was used to fit the models, while the validation set (800 images) was employed to monitor performance and fine-tune hyperparameters. After training was completed, the test set (1000 images) was utilized for the final evaluation of the model on unseen data.

\begin{table}[ht]
\begin{adjustwidth}{-0.3in}{0in}
\caption{Dataset distribution for training, validation, and test sets}
\begin{tabular}{|c|c|c|c|}
\hline
\textbf{Dataset Split} & \textbf{Number of Images} & \textbf{Percentage (\%)} & \textbf{Description} \\ \hline
Training Set           & 3200                     & 64\%                     & Used for model training \\ \hline
Validation Set         & 800                      & 16\%                     & Used for performance monitoring \\ \hline
Test Set               & 1000                     & 20\%                     & Used for final model evaluation \\ \hline
\textbf{Total}         & \textbf{5000}            & \textbf{100\%}            & Complete dataset \\ \hline
\end{tabular}
\end{adjustwidth}
\end{table}


\subsection{Model Training}

This subsection describes the applied vision transformer-based deep learning models with their model architectures, open-set learning implementation and corresponding performance metrics. All experiments were conducted on Google Colab Pro, utilizing the NVIDIA L4 GPU for training, which offers a balanced performance for large models like Swin Transformers. The environment ran on Ubuntu Linux, providing a stable platform for deep learning tasks. We used Python along with several key libraries: TensorFlow and Keras for model training and evaluation, PyTorch and Hugging Face Transformers for Vision Transformer (ViT) models, and Timm for managing advanced transformer architectures like Swin-B and MobileViT. This combination allowed for efficient model implementation and optimization within the deep learning workflow.

\subsubsection{\textbf{ResNet50}}
ResNet50 is a deep convolutional neural network with 50 layers, leveraging residual learning to mitigate the vanishing gradient problem by using shortcut connections~\cite{he2016deep}. The model was fine-tuned using pre-trained ImageNet weights. A custom classification head was added, consisting of a dense layer with 512 units, ReLU activation, and a dropout layer to prevent overfitting. The final output layer employed softmax activation for classifying mosquito species. The model was trained using the Adam optimizer with a learning rate of 0.0001 and sparse categorical cross-entropy loss over 20 epochs. Hyperparameter optimization was conducted to improve the model's performance manually before the model training began, based on best practices and fixed values.

\subsubsection{\textbf{Xception}}
Xception is an extension of the Inception architecture, replacing the standard Inception modules with depthwise separable convolutions, which enhances model efficiency and performance~\cite{chollet2017xception}. This model was fine-tuned using pre-trained weights from ImageNet. A new classification head was added similarly to ResNet50, and training followed the same optimization strategy.

\subsubsection{\textbf{Vision Transformer (vit-base-patch16-224)}}
The Vision Transformer (ViT) model utilizes a transformer architecture designed for image classification by treating image patches as sequences, capturing long-range dependencies in the image~\cite{dosovitskiy2020image}. The specific variant used was \texttt{vit\_base\_patch16\_224}, which splits images into 16×16 patches. The model was pre-trained on ImageNet and fine-tuned for mosquito classification. Image preprocessing involved resizing, center-cropping, and augmenting to improve robustness. The Adam optimizer with a learning rate of 0.0001 and cross-entropy loss was used for training.

\subsubsection{\textbf{Swin Transformer (Swin-B and Swin-S)}}
Swin Transformers efficiently handle image data by focusing attention on localized regions, shifting across layers~\cite{kien2023swintransformer, liu2021swin}. This design allows the model to capture both local and global image features. Swin-B and Swin-S models were fine-tuned using pre-trained ImageNet weights and adapted to the mosquito classification task. Both models used the Adam optimizer and cross-entropy loss. The Swin-S model was trained for five epochs, while Swin-B was trained for ten epochs, with a batch size of 32.

\begin{figure}[h]
  \centering
  \includegraphics[width=1\linewidth]{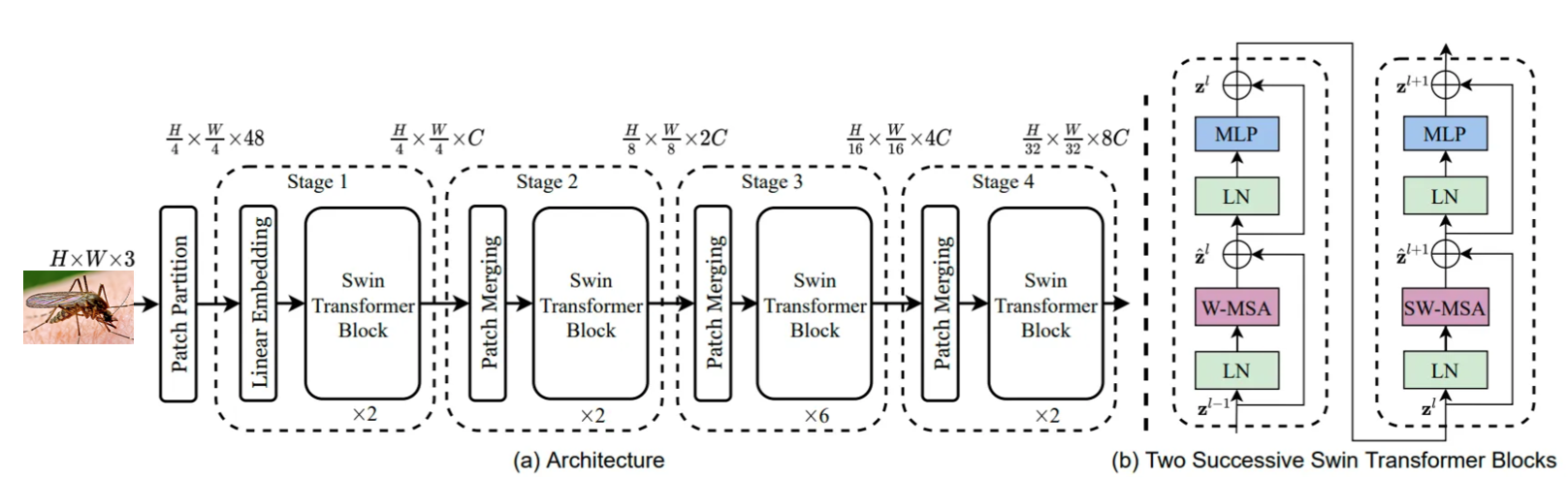}
  \caption{Architecture and Blocks of the Swin Transformer.}
  \label{fig:swin}
\end{figure}

The architecture of the Swin Transformer used in this work is illustrated in Figure~\ref{fig:swin}.

\subsubsection{\textbf{Convolutional Vision Transformer (CVT-13)}}
The Convolutional Vision Transformer (CVT) integrates convolutional layers with transformer components, enabling the model to effectively capture both local features and global context~\cite{wu2021cvt}. Data augmentation was applied to enhance variability and robustness during training. The model was pre-trained on ImageNet and fine-tuned for mosquito classification. The training images were resized, center-cropped, and augmented to maintain consistency, while the CVT-13 model was trained using similar optimization strategies as other models.

\begin{figure}[h]
  \centering
  \includegraphics[width=0.9\linewidth]{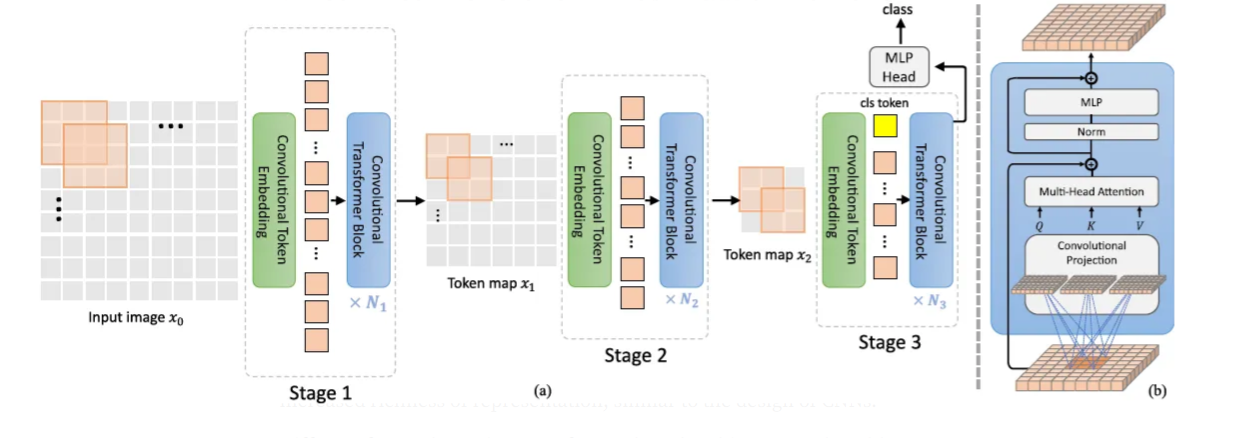}
  \caption{Architecture of CVT-13 technique employed in this work.}
  \label{fig:cvt13}
\end{figure}

The architecture of the CVT-13 model employed in this study is depicted in Figure~\ref{fig:cvt13}.

\subsubsection{\textbf{MobileViT}}
MobileViT combines the efficiency of MobileNets with the representational power of Vision Transformers, making it ideal for mobile and resource-constrained environments~\cite{mehta2110mobilevit}. This lightweight model was pre-trained and fine-tuned for mosquito species classification. Image preprocessing included resizing, center-cropping, and applying random augmentations to improve generalization. The MobileViT model was trained using the Adam optimizer with a learning rate of 0.0001 and cross-entropy loss for ten epochs with a batch size of 32.

\begin{figure}[h]
  \centering
  \includegraphics[width=0.9\linewidth]{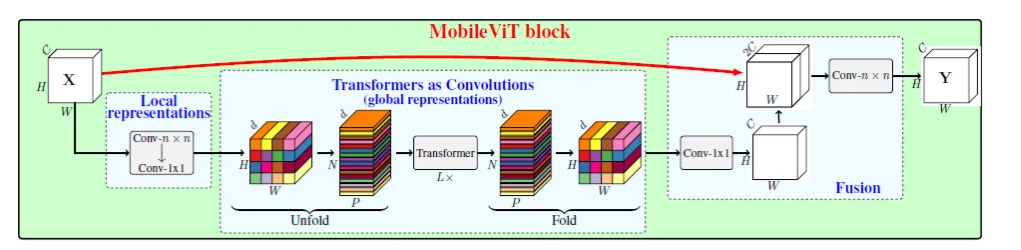}
  \caption{Architecture of MobileViT model applied in this work.}
  \label{fig:mvt}
\end{figure}

The architecture of the MobileViT model applied in this work is shown in Figure~\ref{fig:mvt}.

\subsubsection{Evaluation Metrics}

To comprehensively assess the performance of the trained models, several evaluation metrics were employed, including accuracy, precision, recall, F1-score, Receiver Operating Characteristic (ROC) curve, and confusion matrix. These metrics provide a detailed understanding of how well the models perform on different aspects of classification, such as overall correctness (accuracy), the ability to correctly identify positive instances (precision), the capability to find all relevant instances (recall), and the harmonic mean of precision and recall (F1-score). The ROC curve was used to analyze the trade-off between the true positive rate and the false positive rate at various threshold levels, while the confusion matrix offered insights into the distribution of true positives, true negatives, false positives, and false negatives.
Importantly, all these evaluation metrics were calculated using the test set consisting of 1000 images, which was completely held out during the model training and validation phases. This ensures that the evaluation results reflect the models' performance on unseen data, providing a robust measure of their generalization capabilities. 

\subsection{Open-set Learning Implementation}
\noindent  The dataset for open-set learning was prepared by constructing a new extended test set comprising a new class "unknown" with the existing 10 classes. This new class has 100 unknown images that were not seen during the training phase. This test set includes images of bees (33 images), butterflies (34 images), and flies (33 images). To utilize the trained model, we calculated the Mean Activation Vector (MAV) for the known classes and fitted the Weibull distribution to model the tails of these activations. Finally, a threshold for the OpenMax layer was established based on the Weibull distribution, which helps in determining the likelihood of a sample belonging to an unknown class.
\subsubsection{Logits Extraction:}
For each image $\mathbf{x}_i$ in the dataset, the model outputs logits $\mathbf{z}_i$~\cite{wei2022mitigating, shao2023deep} expressed as:

\begin{equation}
\label{logits}
\mathbf{z}_i = \text{model}(\mathbf{x}_i)
\end{equation}

where $\mathbf{z}_i \in \mathbb{R}^C$ and $C$ is the number of classes ($C$ = 10).

\subsubsection{Mean Activation Vector (MAV):}
For each class $C$, the MAV coefficient ~\cite{neal2018open} has been calculated from the respective logits in \ref{logits} as:

\begin{equation}
\label{MAV}
\text{MAV}_c = \frac{1}{N_c} \sum_{i=1}^{N_c} \mathbf{z}_{i,c}
\end{equation}

where $\mathbf{z}_{i,c}$ are the logits of images that belong to class $C$ and $N_c$ is the number of such images.

\subsubsection{Euclidean Distances:}
The Euclidean distance ~\cite{giusti2022proportional} between each logit vector and its associated class MAV is computed as follows:

\begin{equation}
\label{euclidean}
d_{i,c} = \| \mathbf{z}_{i,c} - \text{MAV}_c \|_2
\end{equation}

\subsubsection{Weibull Distribution Fitting:}
A Weibull distribution to the distances for each class $C$ is approximated as:

\begin{equation}
\label{Weibull}
\text{CDF}_{\text{Weibull}}(a; b, \gamma, \mu) = 1 - \exp \left( - \left( \frac{a - \mu}{\gamma} \right)^b \right)
\end{equation}

where $b$, $\gamma$ and $\mu$ are the shape, scale and location parameters, respectively.

\subsubsection{Recalibrated Scores:}

For each logit vector, the recalibrated scores ~\cite{ge2017generative} are obtained using the Weibull CDF as:

\begin{equation}
\label{Recalibrated}
r_c = 1 - \text{CDF}_{\text{Weibull}}(d_{i,c})
\end{equation}

\subsubsection{OpenMax Threshold:}
If the maximum recalibrated score, defined in \ref{Recalibrated}, is below a definite threshold, the inference sample is classified as ``Unknown.'' If not, the class is assigned with the highest recalibrated score ~\cite{wang2022openauc, zhou2021learning}:

\begin{equation}
\label{OpenMax}
\text{prediction} = 
\begin{cases} 
C + 1 & \text{if } \max(r_c) < \text{threshold} \\
\arg \max(r_c) & \text{otherwise}
\end{cases}
\end{equation}

The proposed open-set learning algorithm based on the OpenMax probability estimation applied in this work has been summarized in Algorithm \ref{alg:openmax_modified}.

\begin{algorithm}
\caption{OpenMax Probability Calculation with Exclusion of Unknown or Ambiguous Inputs}
\label{alg:openmax_modified}
\begin{algorithmic}[1]
\State \textbf{Input:} Activation vector $\mathbf{a}(\mathbf{z}) = [a_1(\mathbf{z}), \ldots, a_M(\mathbf{z})]$
\State \textbf{Input:} Mean vectors $\nu_k$ and Weibull parameters $\psi_k = (\tau_k, \lambda_k, \kappa_k)$
\State \textbf{Input:} $\beta$, the number of leading classes to modify
\Statex
\Procedure{OpenMax}{$\mathbf{a}, \nu, \psi, \beta$}
    \State $\mathbf{r}(\mathbf{z}) \gets \text{argsort}(\mathbf{a}(\mathbf{z}))$
    \State $\theta_k \gets 1 \text{ for all } k$
    \For{$j = 1$ to $\beta$}
        \State $\theta_{r(j)}(\mathbf{z}) \gets 1 - \frac{\beta - j}{\beta} \cdot e^{\left( \frac{\|\mathbf{a}_{r(j)}(\mathbf{z}) - \nu_{r(j)}\|_2}{\lambda_{r(j)}} \right)^{\kappa_{r(j)}}}$
    \EndFor
    \State $\tilde{\mathbf{a}}(\mathbf{z}) \gets \mathbf{a}(\mathbf{z}) \odot \theta(\mathbf{z})$
    \State $\tilde{a}_0(\mathbf{z}) \gets \sum_j a_j(\mathbf{z})(1 - \theta_j(\mathbf{z}))$
    \State $\tilde{P}(c = k|\mathbf{z}) \gets \frac{e^{\tilde{a}_k(\mathbf{z})}}{\sum_{j=0}^{M} e^{\tilde{a}_j(\mathbf{z})}}$
    \State $c^* \gets \arg \max_k \tilde{P}(c = k|\mathbf{z})$
    \If{$c^* = 0$ or $\tilde{P}(c = c^*|\mathbf{z}) < \epsilon$}
        \State Reject input
    \EndIf
\EndProcedure
\end{algorithmic}
\end{algorithm}

The OpenMax algorithm processes input by accepting an activation vector $\mathbf{a}(\mathbf{z})$, mean vectors $\nu_k$, and Weibull parameters $\psi_k$, along with $\beta$, the number of leading classes to be adjusted. The algorithm starts by sorting $\mathbf{a}(\mathbf{z})$ to identify the classes with the highest activations and initializes scaling factors $\theta_k$ to 1 for all classes. For each of the top $\beta$ classes, it modifies the scaling factors $\theta_{r(j)}$ based on the Euclidean distance between the class activation and its mean vector, adjusted by the Weibull shape and scale parameters. This scaling reduces the influence of activations that are far from the model's training data, representing increased uncertainty. The modified activations $\tilde{\mathbf{a}}(\mathbf{z})$ are then computed by applying these scaling factors, and an additional activation $\tilde{a}_0(\mathbf{z})$ is calculated for the unknown class, representing all reduced probabilities. Softmax is applied to these activations to compute the final class probabilities $\tilde{P}(c = k|\mathbf{z})$. The algorithm decides to reject the input if the predicted class is the unknown class or if its probability is lower than a set threshold $\epsilon$, enhancing the model's ability to deal with unknown or ambiguous inputs effectively.

\subsection{Top-level overview of the proposed mosquito classification system.}

\begin{figure}
    \centering
    \includegraphics[width=1\linewidth]{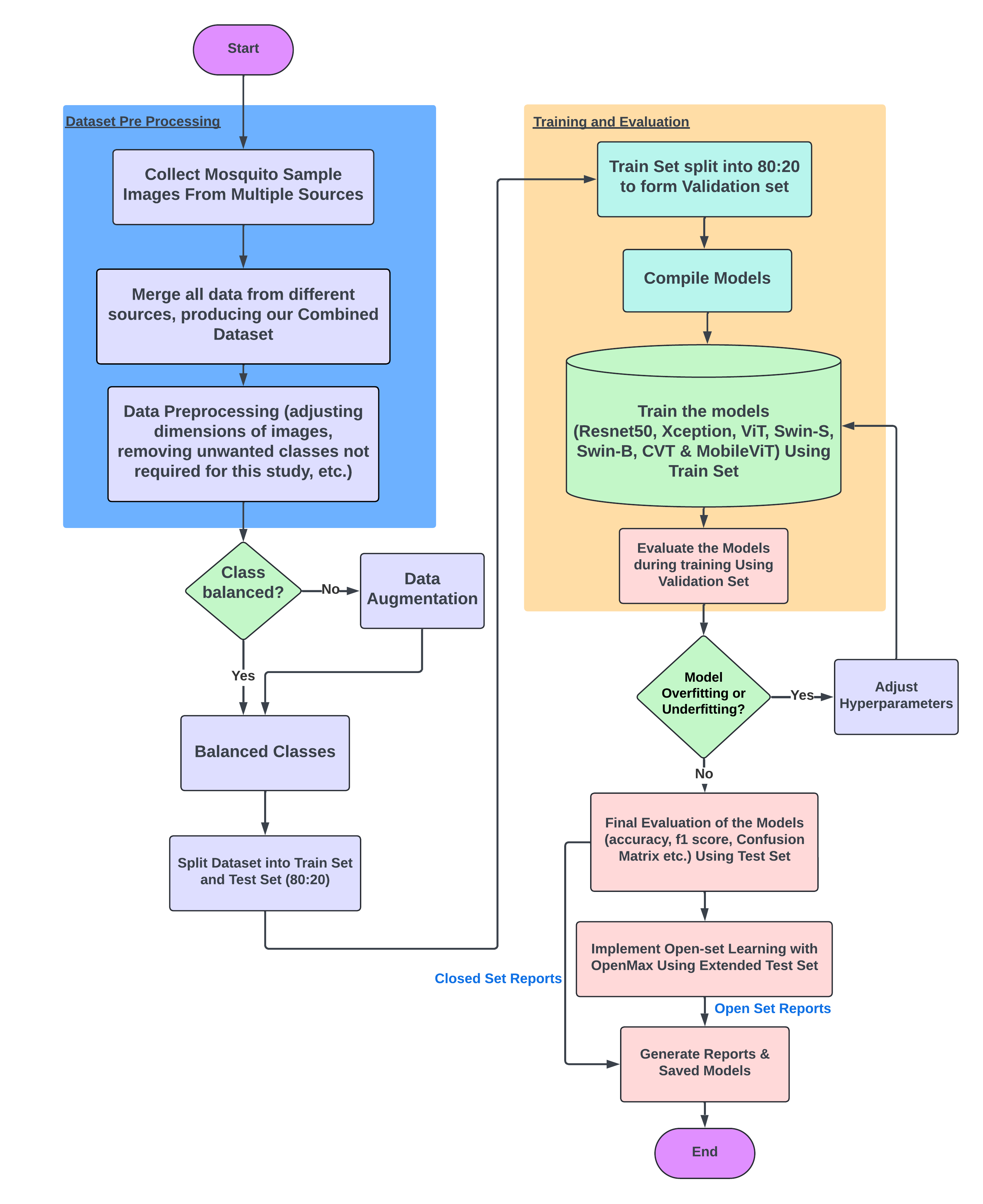}
    \caption{A detailed workflow diagram of the proposed mosquito classification system.}
    \label{fig:workflow_diagram}
\end{figure}

Figure~\ref{fig:workflow_diagram} illustrates the workflow for the proposed deep learning-based mosquito classification system. The process begins with the collection and merging of various mosquito image datasets. After the data is collected, it is split and augmented to increase variability and dataset size. At a decision point, the workflow checks if the data is balanced. If the data is unbalanced, balancing techniques are applied to ensure even class distribution. Once the data is balanced, various models, including ResNet50, Xception, Vision Transformer (ViT), Swin-S, Swin-B, Convolutional Vision Transformer (CVT), and MobileViT, are compiled and trained. The next step in the workflow is checking if the models have achieved the desired accuracy. If the performance does not meet the expectations, hyperparameters are adjusted, and the models are retrained. Once the desired accuracy is achieved, the workflow proceeds to implement open-set learning using the OpenMax layer and the Weibull distribution to handle unknown classes during inference. The final step in the process involves generating classification reports based on the trained model's performance.

\section{Results and Discussion}
This section represents the findings of this study for both closed and open-set learning for static and non-stationary environments, respectively. The first subsection contains mosquito image classification for closed-set. All the evaluations of the model's performance were performed using the unseen data during training or test set. Next, the open-set learning results are illustrated for the applied deep learning techniques employing the OpenMax dataset and Weibull distribution using the extended test set data, specially prepared for this particular task. Finally, the proposed mosquito classification system has been compared with the related articles.

\subsection{Results for Closed-set Learning}
Table \ref{resultsintro} presents the performance metrics for various models applied to closed-set learning in the proposed mosquito classification. The Swin-B Transformer achieves the highest overall accuracy at 99.60\%, with corresponding high precision (0.994), recall (0.996), and F1 score (0.996), indicating its superior performance. The MobileViT model also performs well, achieving a high accuracy of 98.90\% and an F1 score of 0.989, while utilizing significantly fewer parameters (5.58 million), making it an efficient model for mosquito classification. The Xception model follows closely with an accuracy of 99.20\% and an F1 score of 0.991. On the other hand, ResNet50 shows the lowest accuracy at 94.50\%, but still demonstrates reliable performance with a precision of 0.937 and a recall of 0.933. All the results were based on the fine-tuned model's performance on unseen data during the training phase, the test set. These results highlight the effectiveness of transformer-based models, particularly Swin-B and MobileViT, in accurately classifying mosquito species while maintaining computational efficiency.


\begin{table}[htbp]
\caption{Performance metrics of the applied models for closed-set learning}
\label{resultsintro}
\centering
\begin{adjustbox}{max width=\textwidth}
\begin{tabular}{cccccc}
\hline
Model      &       Precision &  Recall &     F1 score  &  Accuracy ($\%$) &      Parameters (million) \\
\hline  
ResNet50       &  0.937 &  0.933  &  0.933 & 94.50 &  23.59      \\
ViT Base       &  0.979  & 0.978 &  0.977 & 97.80  &  85.81  \\
CvT-13     &  0.985 & 0.984 &  0.984 & 98.40 &  19.62   \\
MobileViT      &  0.989  & 0.989 & 0.989 & 98.90 &   5.58    \\
Xception       & 0.991  & 0.911 & 0.991  & 99.20 &  22.91   \\
\textbf{Swin-B}    & \textbf{ 0.996}  & \textbf{0.996}  & \textbf{ 0.996}  & \textbf{99.60}  &   86.75 \\
Swin-S      & 0.983  & 0.981 & 0.981  & 98.25 &  48.84 \\
\hline
\end{tabular}
\end{adjustbox}
\end{table}

\begin{figure}[htbp]
  \centering
  \begin{subfigure}[b]{0.45\textwidth}
    \includegraphics[width=\textwidth]{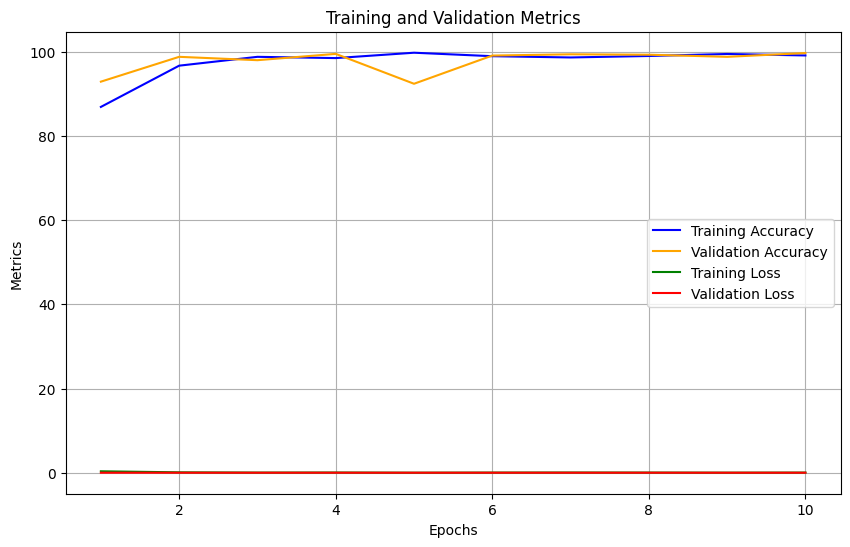}
    \caption{Swin-B}
    \label{gsubfig1}
  \end{subfigure}
  \hfill
  \begin{subfigure}[b]{0.45\textwidth}
    \includegraphics[width=\textwidth]{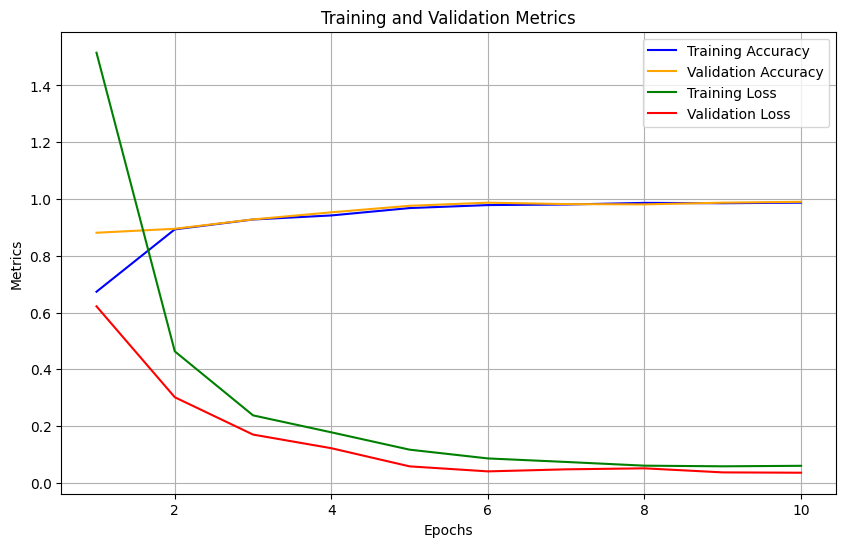}
    \caption{MobileViT}
    \label{gsubfig2}
  \end{subfigure}
  \vskip\baselineskip
  \begin{subfigure}[b]{0.45\textwidth}
    \includegraphics[width=\textwidth]{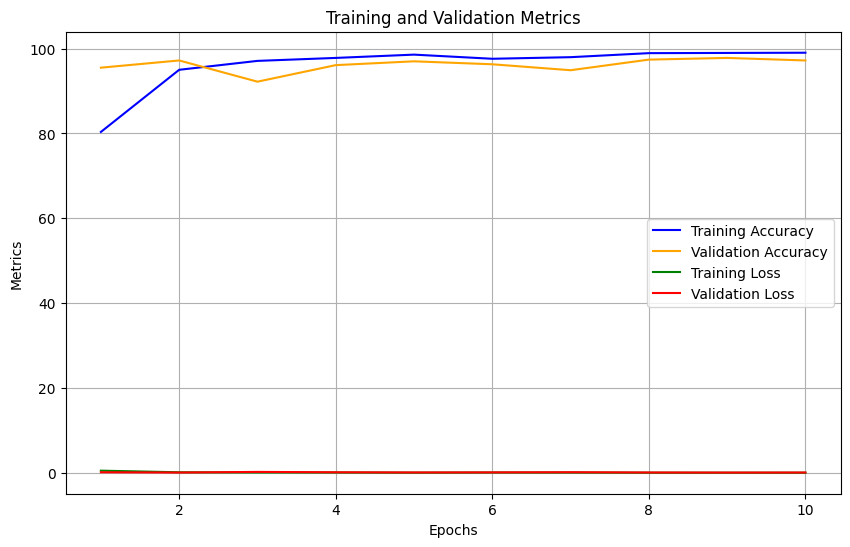}
    \caption{ViT Base}
    \label{gsubfig3}
  \end{subfigure}
  \hfill
  \begin{subfigure}[b]{0.45\textwidth}
    \includegraphics[width=\textwidth]{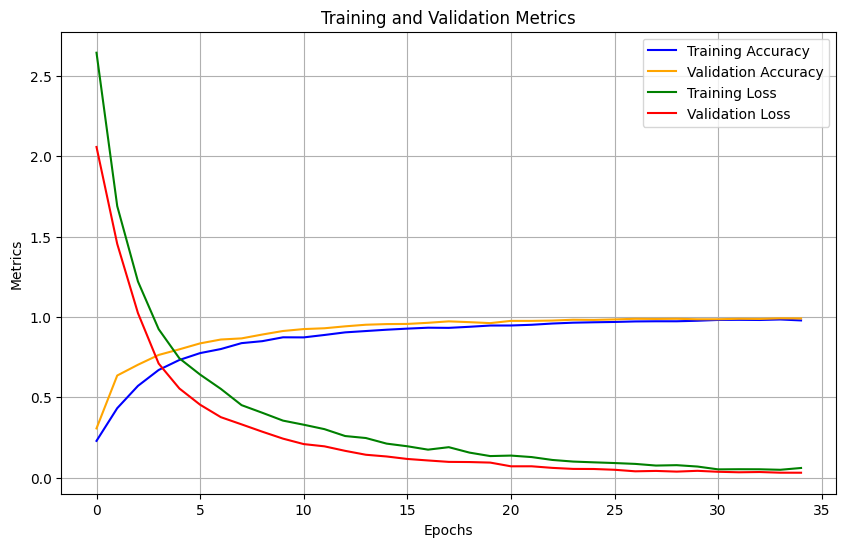}
    \caption{Xception}
    \label{gsubfig4}
  \end{subfigure}
  \caption{Training and validation accuracies and losses vs. epochs of the applied models. }
  \label{fig:graph_2}
\end{figure}

Figure~\ref{fig:graph_2} shows the training and validation accuracies and losses vs. epochs for the applied models with closed-set learning. The model demonstrates a consistent rise in training accuracy, with validation accuracy improving, though with some fluctuations. The loss metrics indicate a steady reduction in training and validation loss, highlighting the model’s capability to learn from the data and generalize well. 
In Figure~\ref{gsubfig2}, the MobileViT model shows an upward trend in training and validation accuracy, though the validation accuracy increases gradually. Both training and validation loss decrease, but the validation loss plateaus slightly, indicating potential challenges in further improving validation performance. For the ViT Base model, Figure~\ref{gsubfig3} illustrates an increasing trend in training and validation accuracy, reflecting effective training. The training loss decreases consistently, while the validation loss reduces, albeit with minor fluctuations, suggesting good generalization with occasional overfitting. According to Figure~\ref{gsubfig4}, the Xception model indicates a steady rise in training accuracy, with the validation accuracy showing a similar pattern but with slight instability. The training loss decreases significantly, and the validation loss also reduces, though it exhibits minor variations, indicating some challenges in validation performance consistency.

\begin{figure}[htbp]
  \centering
  \begin{subfigure}[b]{0.45\textwidth}
    \includegraphics[width=\textwidth]{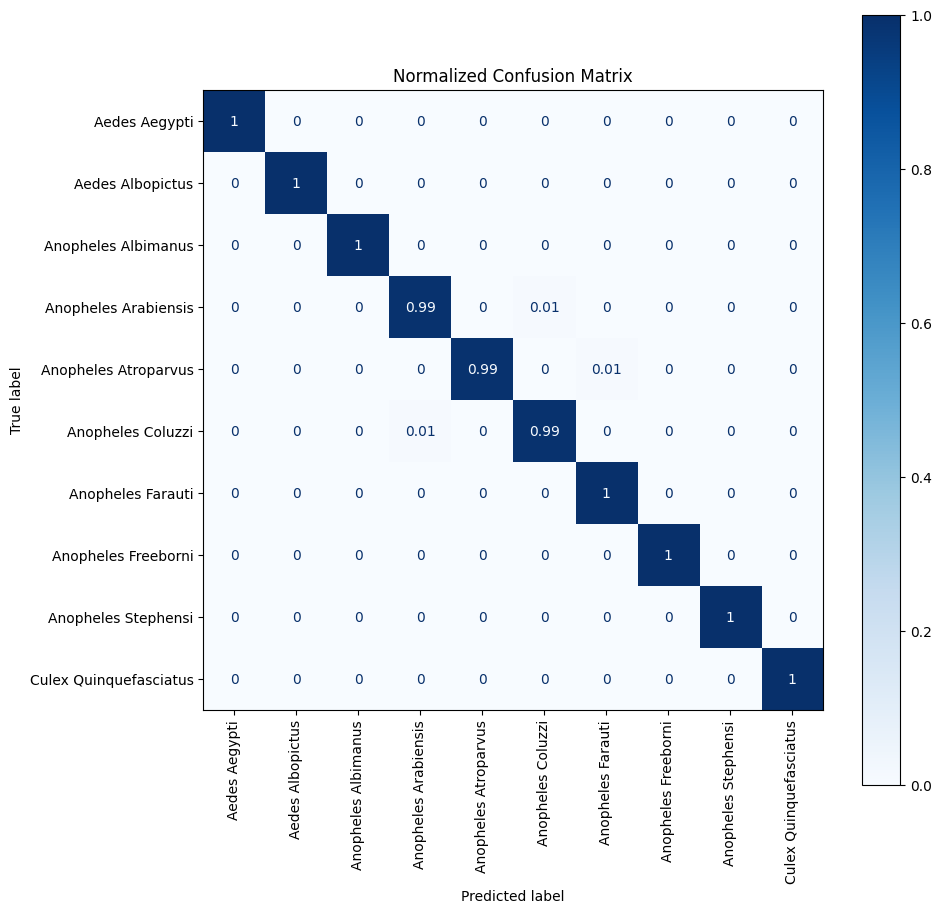}
    \caption{Swin-B}
    \label{subfig1}
  \end{subfigure}
  \hfill
  \begin{subfigure}[b]{0.45\textwidth}
    \includegraphics[width=\textwidth]{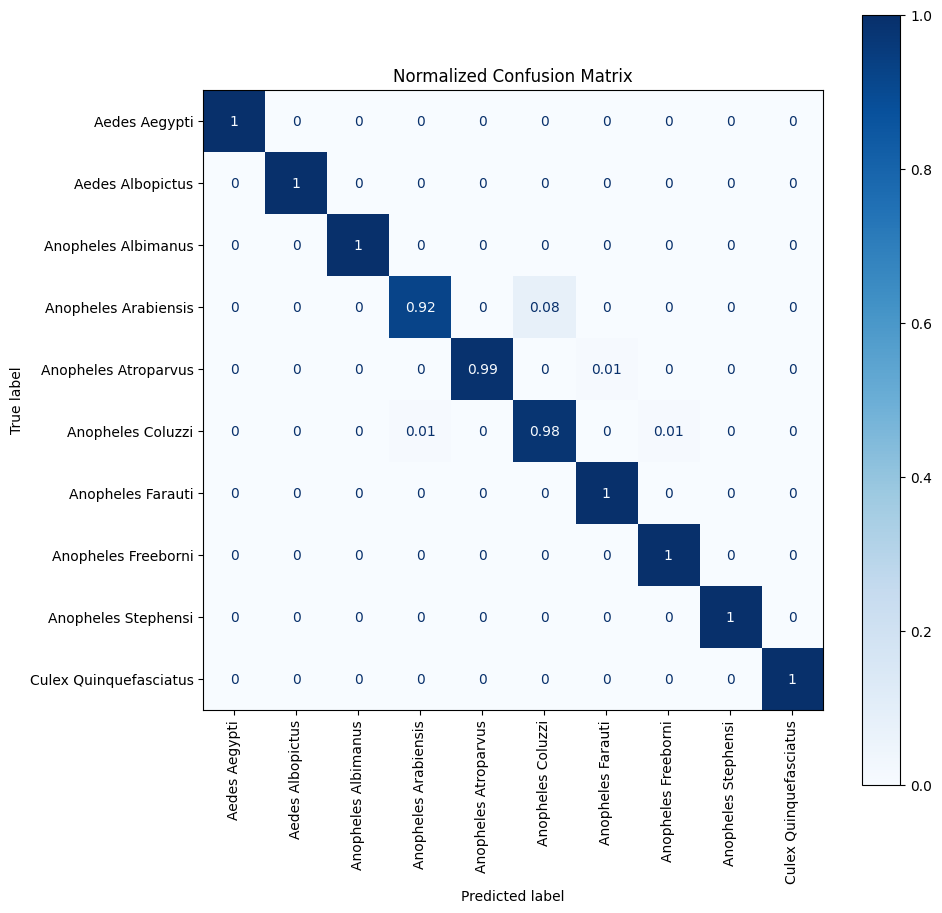}
    \caption{MobileViT}
    \label{subfig2}
  \end{subfigure}
  \vskip\baselineskip
  \begin{subfigure}[b]{0.45\textwidth}
    \includegraphics[width=\textwidth]{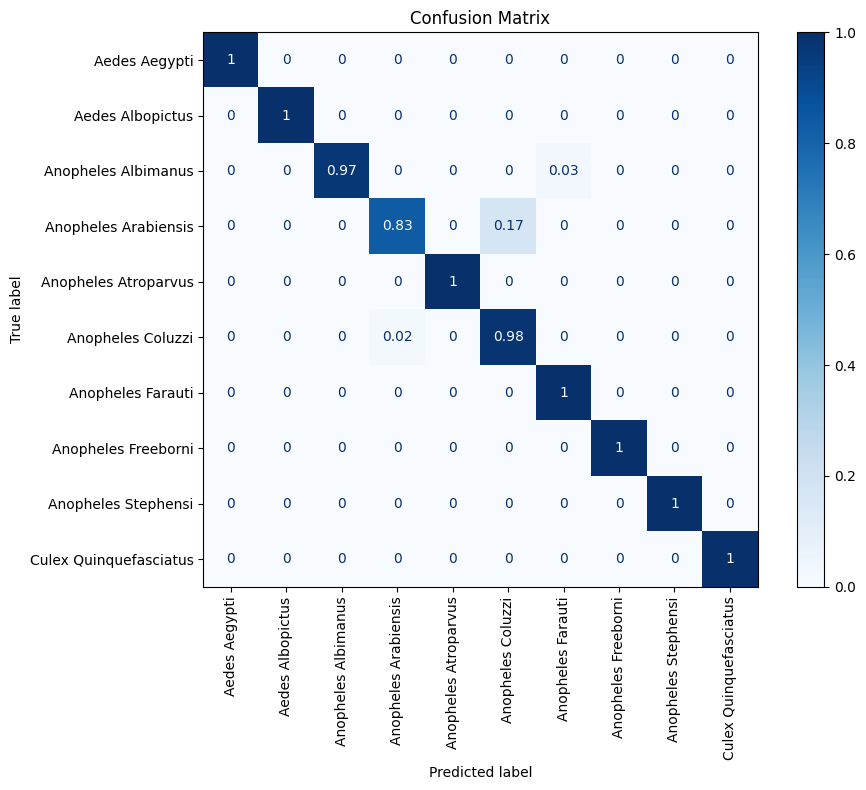}
    \caption{ViT Base}
    \label{subfig3}
  \end{subfigure}
  \hfill
  \begin{subfigure}[b]{0.45\textwidth}
    \includegraphics[width=\textwidth]{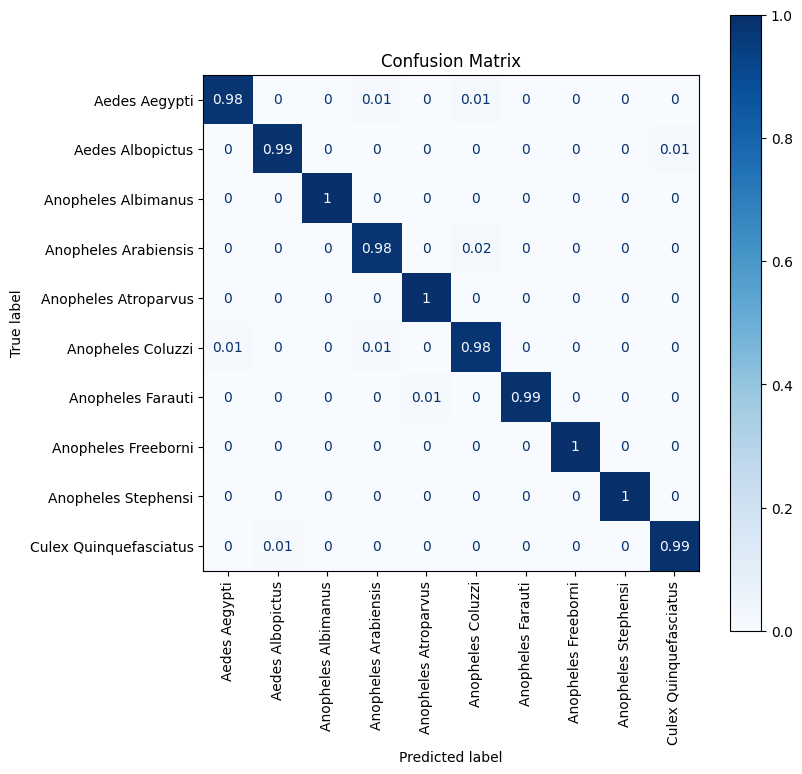}
    \caption{Xception}
    \label{subfig4}
  \end{subfigure}
  \caption{Normalized confusion matrices of the applied models using the test set}.
  \label{fig:subfigures_part2}
\end{figure}

Figure~\ref{fig:subfigures_part2} represents the normalized confusion matrix of the applied models for the test set samples of the ten mosquito classes (unseen data during the training phase). According to Figure~\ref{subfig1}, the Swin Transformer model demonstrates high accuracy, with most species correctly classified. However, there are minor misclassifications, such as Anopheles arabiensis being confused with Anopheles Coluzzii and Anopheles Atroparvus. The MobileViT model attained almost similar performance with Swin in mosquito classification. This lightweight model with significantly reduced parameters achieves high classification accuracy overall, although some misclassifications are observed, such as Anopheles Arabiensis being occasionally misidentified as Anopheles coluzzii. The ViT Transformer model shows exceptional performance, though some misclassifications, such as Anopheles arabiensis being occasionally misidentified as Anopheles coluzzii and Anopheles albimanus.

\begin{figure}[htbp]
  \centering
  \begin{subfigure}[b]{0.45\textwidth}
    \includegraphics[width=\textwidth]{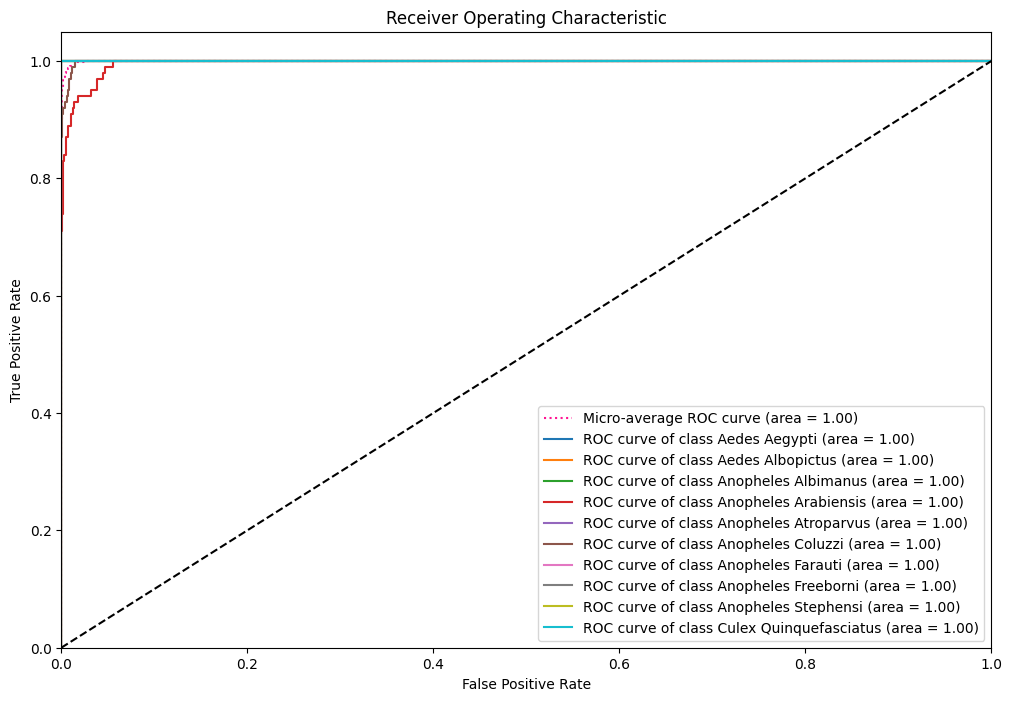}
    \caption{Swin-B}
    \label{swingsubfig1}
  \end{subfigure}
  \hfill
  \begin{subfigure}[b]{0.45\textwidth}
    \includegraphics[width=\textwidth]{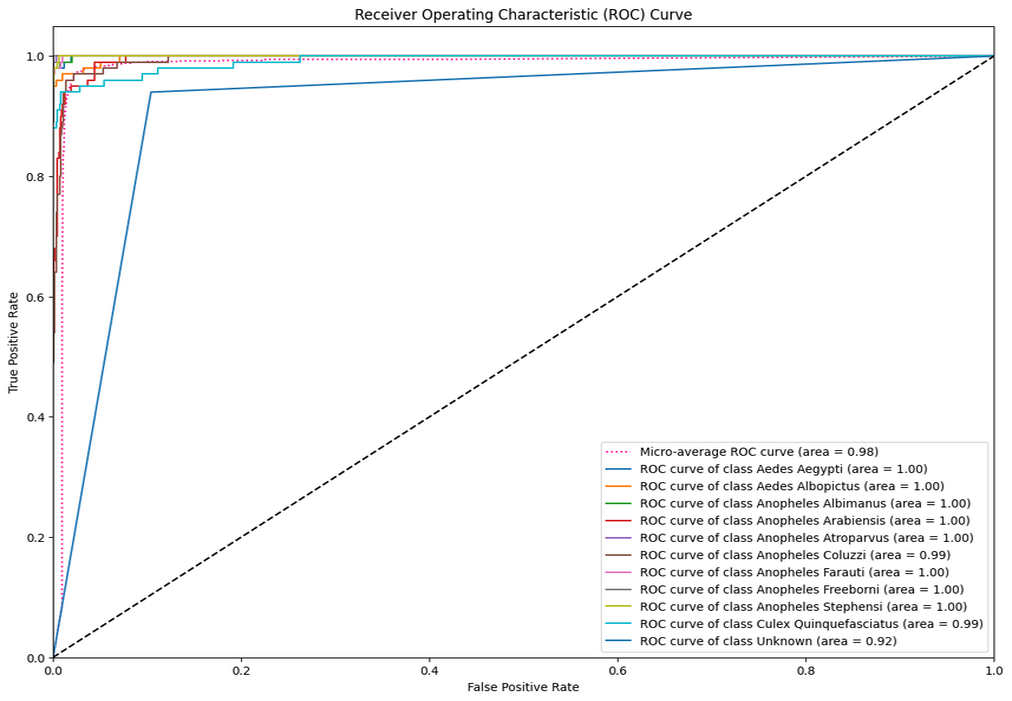}
    \caption{Xception}
    \label{swingsubfig3}
  \end{subfigure}
  \hfill
  \begin{subfigure}[b]{0.45\textwidth}
    \includegraphics[width=\textwidth]{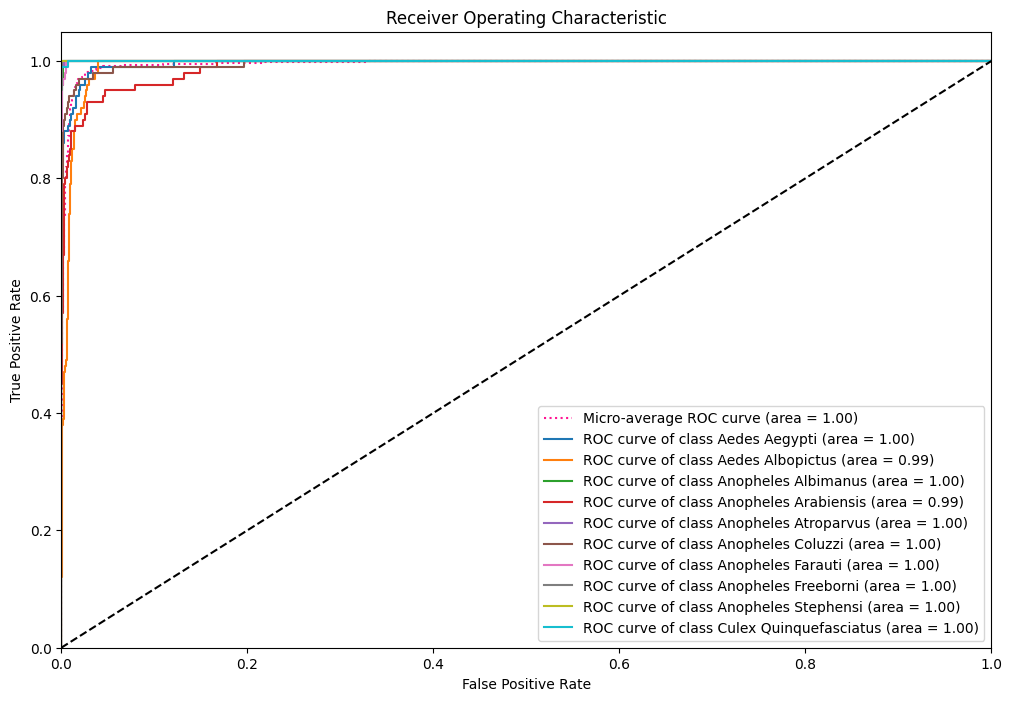}
    \caption{ViT}
    \label{swingsubfig5}
  \end{subfigure}
  \hfill
  \begin{subfigure}[b]{0.45\textwidth}
    \includegraphics[width=\textwidth]{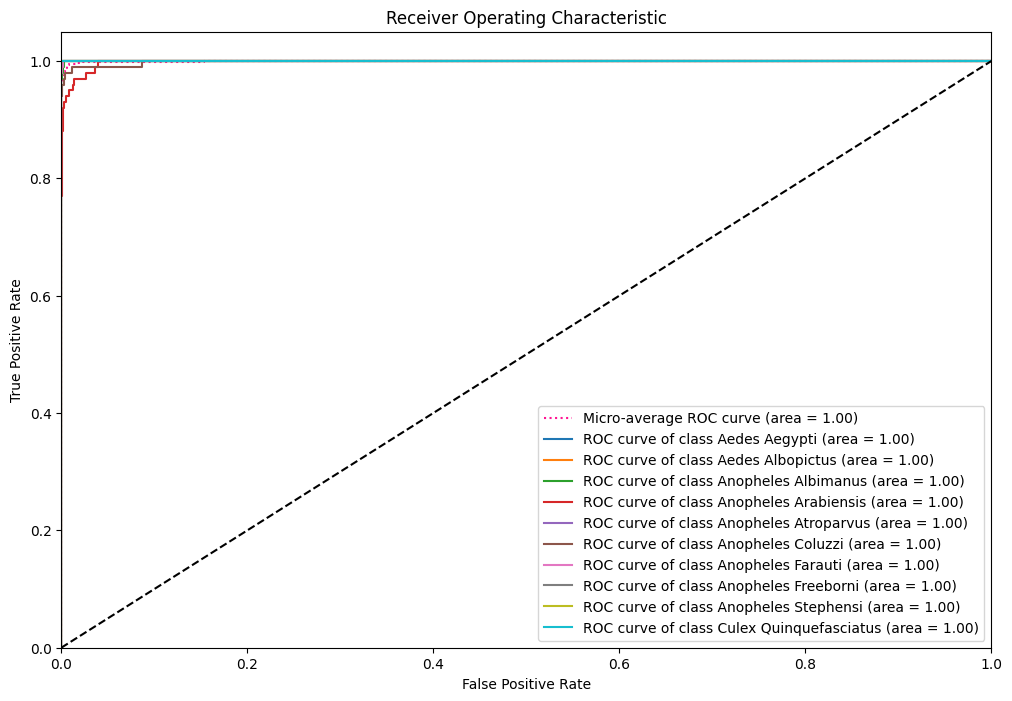}
    \caption{Swin-S}
    \label{swingsubfig7}
  \end{subfigure}
  \caption{ROC curves of different models for closed-set learning using the test set}.
  \label{fig:swingraph_2}
\end{figure}

Figure~\ref{fig:swingraph_2} illustrates the ROC curves for different models applied to closed-set learning, including Swin-B, Xception, ViT, and Swin-S. The curves demonstrate the high performance of the applied models, with most achieving near-perfect AUC scores, indicating their strong capability in distinguishing between mosquito species.

\begin{figure}[htbp]
  \centering
  \begin{subfigure}{0.19\textwidth}
    \centering
    \includegraphics[width=\linewidth]{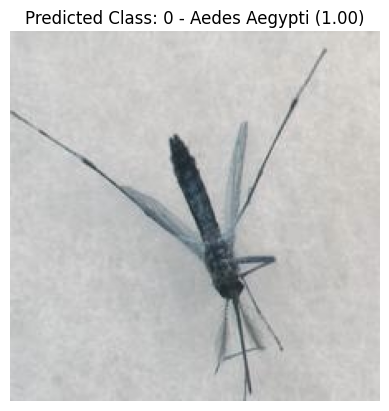}
    \caption{Ae. Aegypti}
  \end{subfigure}
  \begin{subfigure}{0.19\textwidth}
    \centering
    \includegraphics[width=\linewidth]{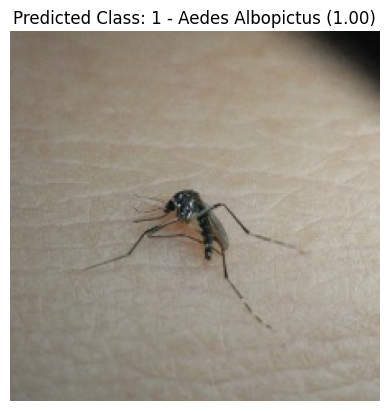}
    \caption{Ae. Albopictus}
  \end{subfigure}
  \begin{subfigure}{0.19\textwidth}
    \centering
    \includegraphics[width=\linewidth]{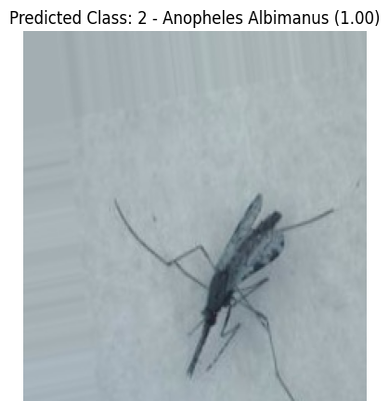}
    \caption{An. Albimanus}
  \end{subfigure}
  \begin{subfigure}{0.19\textwidth}
    \centering
    \includegraphics[width=\linewidth]{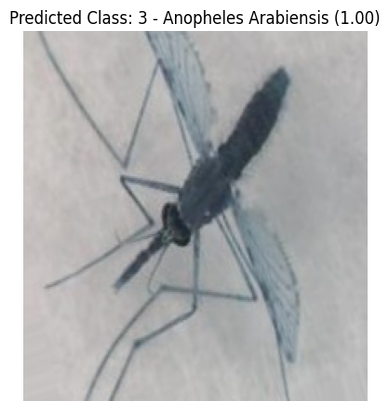}
    \caption{An. Arabiensis }
  \end{subfigure}
  \begin{subfigure}{0.19\textwidth}
    \centering
    \includegraphics[width=\linewidth]{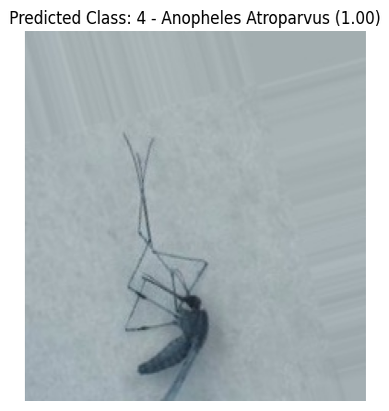}
    \caption{An. Atroparvus}
  \end{subfigure}
  
  \begin{subfigure}{0.19\textwidth}
    \centering
    \includegraphics[width=\linewidth]{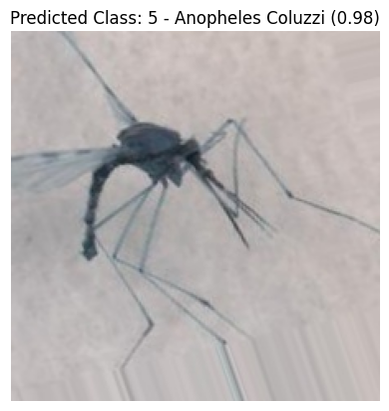}
    \caption{An. Coluzzi}
  \end{subfigure}
  \begin{subfigure}{0.19\textwidth}
    \centering
    \includegraphics[width=\linewidth]{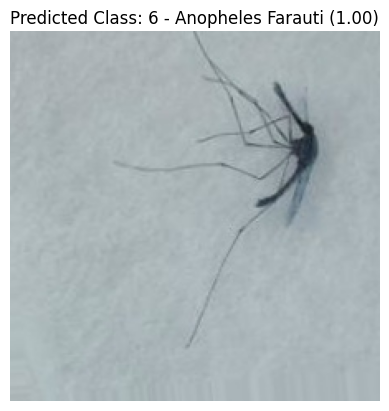}
    \caption{An. Farauti}
  \end{subfigure}
  \begin{subfigure}{0.19\textwidth}
    \centering
    \includegraphics[width=\linewidth]{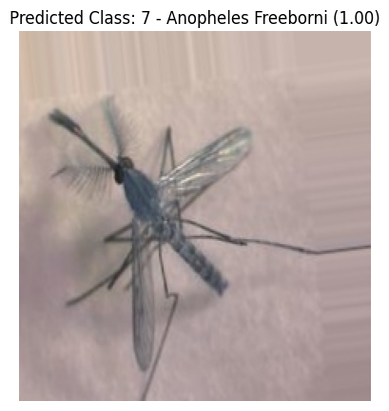}
    \caption{An. Freeborni}
  \end{subfigure}
  \begin{subfigure}{0.19\textwidth}
    \centering
    \includegraphics[width=\linewidth]{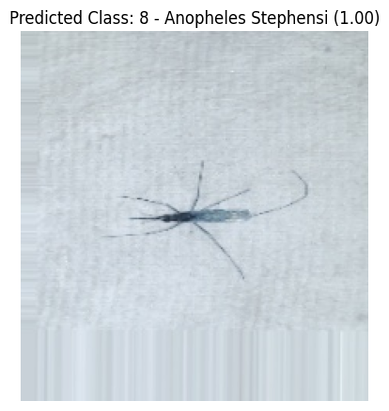}
    \caption{An. Stephensi}
  \end{subfigure}
  \begin{subfigure}{0.19\textwidth}
    \centering
    \includegraphics[width=\linewidth]{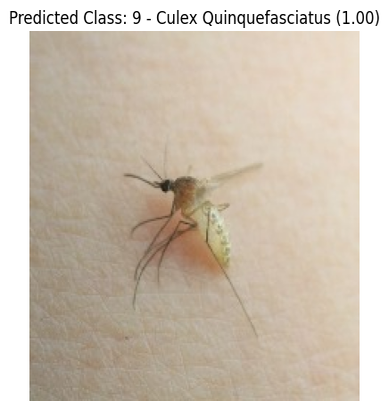}
    \caption{Cx. Quinquefasciatus}
  \end{subfigure}
  \caption{Classifications of the Swin-B model with actual and predicted classes for closed-set learning on the test set images.}
  \label{fig:swin-BWIOM}
\end{figure}

Figure~\ref{fig:swin-BWIOM} illustrates the classification results of the Swin-B model, displaying the actual and predicted classes for various mosquito species in a closed-set learning scenario. The model accurately identifies each species, with high confidence scores (1.00) for most predictions, demonstrating its effectiveness in distinguishing between different mosquito species.

The Swin-B model, known for its high precision in classifying individual mosquitoes, also excels at detecting and categorizing multiple mosquitoes within a single image. Thanks to its transformer-based architecture, which captures intricate spatial relationships and minute details, the model can effectively isolate and recognize each mosquito, even in complex or cluttered environments. By employing attention mechanisms, Swin-B can focus on various sections of the image, allowing it to detect and classify multiple mosquitoes, despite challenges like overlapping or different poses. This ability makes Swin-B particularly valuable for real-world applications where mosquitoes often appear in groups.

\begin{figure}[ht]
    \centering
    \begin{subfigure}{0.45\textwidth}
        \includegraphics[width=\textwidth]{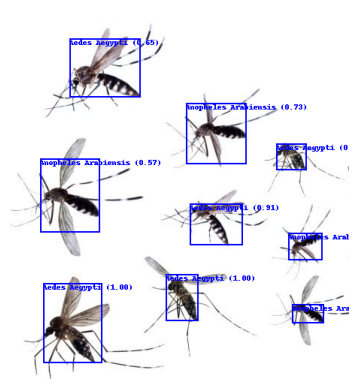}
        \caption{Example of multiple mosquito detection by model Swin-B.}
        \label{fig:image1}
    \end{subfigure}
    \hfill
    \begin{subfigure}{0.45\textwidth}
        \includegraphics[width=\textwidth]{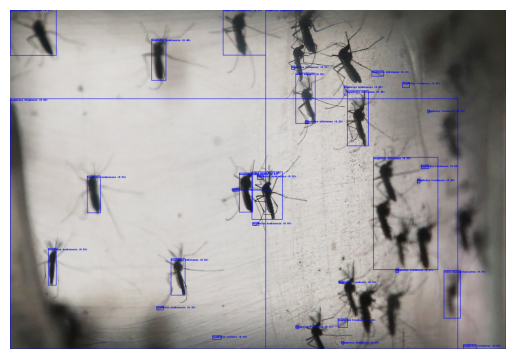}
        \caption{Swin-B model classifying mosquitoes in a cluttered environment.}
        \label{fig:image2}
    \end{subfigure}
    \hfill
    \begin{subfigure}{0.45\textwidth}
        \includegraphics[width=\textwidth]{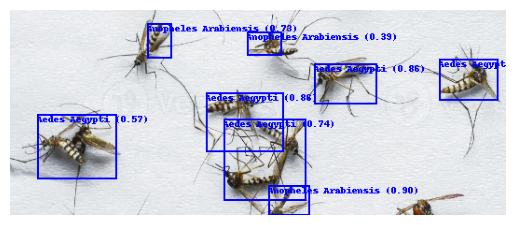}
        \caption{Swin-B model classifying mosquitoes in a cluttered environment.}
        \label{fig:image3}
    \end{subfigure}
    \caption{Detection and classification of mosquitoes using Swin-B model in two different scenarios.}
\end{figure}

\noindent As shown in \figurename~\ref{fig:image1}, \figurename~\ref{fig:image2} and \figurename~\ref{fig:image3}, the model demonstrates its capability to detect mosquitoes accurately in images where multiple mosquitoes are present, identifying each mosquito individually even in crowded scenes. 

\subsection{Open-set Learning Implementation Results}

\begin{figure}[htbp]
  \centering
  \begin{subfigure}{0.19\textwidth}
    \centering
    \includegraphics[width=\linewidth]{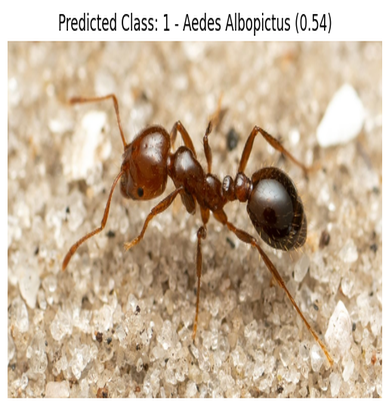}
    \caption{Ant}
    \label{fig:ant}
  \end{subfigure}
  \begin{subfigure}{0.19\textwidth}
    \centering
    \includegraphics[width=\linewidth]{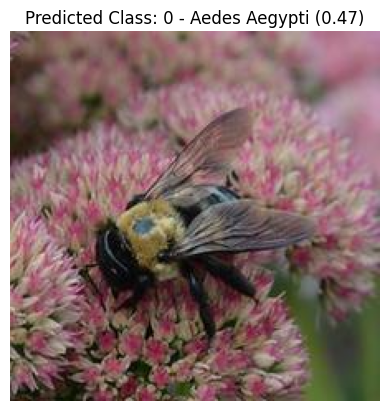}
    \caption{Bee}
    \label{fig:bee1}
  \end{subfigure}
  \begin{subfigure}{0.19\textwidth}
    \centering
    \includegraphics[width=\linewidth]{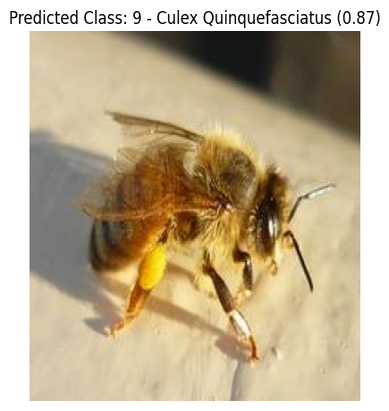}
    \caption{Bee}
    \label{fig:bee2}
  \end{subfigure}
  \begin{subfigure}{0.19\textwidth}
    \centering
    \includegraphics[width=\linewidth]{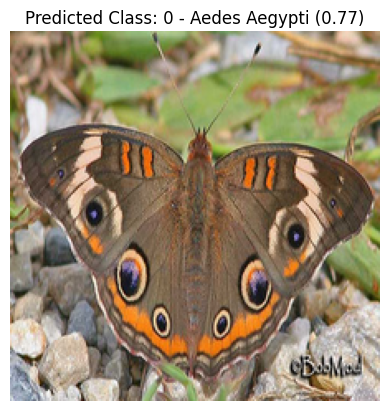}
    \caption{Butterfly}
    \label{fig:butterfly1}
  \end{subfigure}
  \begin{subfigure}{0.19\textwidth}
    \centering
    \includegraphics[width=\linewidth]{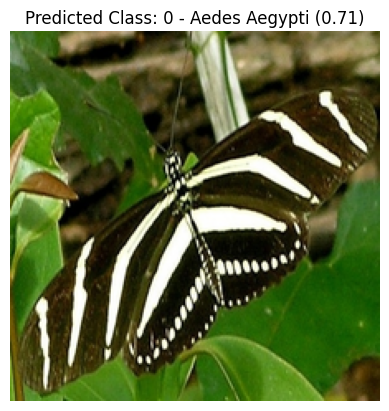}
    \caption{Butterfly}
    \label{fig:butterfly2}
  \end{subfigure}
  \caption{Wrong classification of unknown insects in static setting (prior to implementing open-set learning).}
  \label{fig:WOM}
\end{figure}

The primary challenge we encountered during our research was that our models were susceptible to misclassification when presented with images outside the domain of our dataset. Our models, which were trained exclusively on our mosquito dataset, incorrectly classified images of unknown classes (e.g., ants, bees, butterflies) as one of the ten known mosquito classes. This issue is illustrated in Figure~\ref{fig:WOM}, where even images of ants, bees, and butterflies were erroneously predicted to belong to one of the mosquito classes. Figure~\ref{fig:ant} shows that an ant was classified as Aedes albopictus with a confidence score of 0.54. In figure~\ref{fig:bee1},\ref{fig:bee2}, two different types of bee were classified as Aedes aegypti and Culex quinquefasciatus with scores of 0.47 and 0.87, respectively. In figure~\ref{fig:butterfly1},\ref{fig:butterfly2}, two different types of butterfly were classified as Aedes aegypti with scores of 0.77 and 0.71, respectively. To address this problem, we incorporated the OpenMax method, which allowed us to effectively identify and label outlier images as unknown, thereby improving the robustness of our model in handling images from outside the training dataset.

\begin{figure}[htbp]
  \centering
  \begin{subfigure}{0.19\textwidth}
    \centering
    \includegraphics[width=\linewidth]{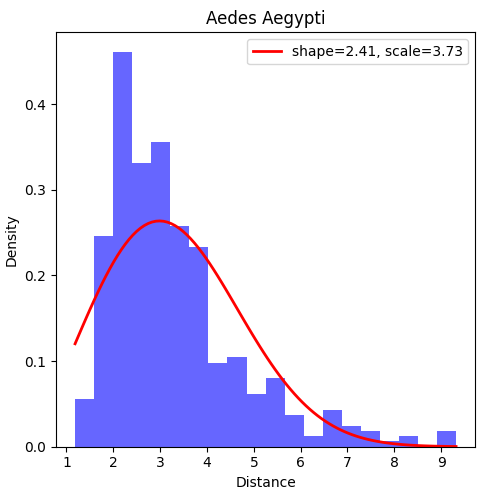}
    \caption{Ae. Aegypti}
  \end{subfigure}
  \begin{subfigure}{0.19\textwidth}
    \centering
    \includegraphics[width=\linewidth]{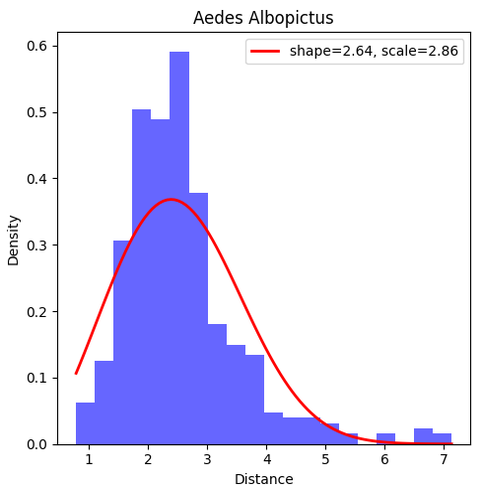}
    \caption{Ae. Albopictus}
  \end{subfigure}
  \begin{subfigure}{0.19\textwidth}
    \centering
    \includegraphics[width=\linewidth]{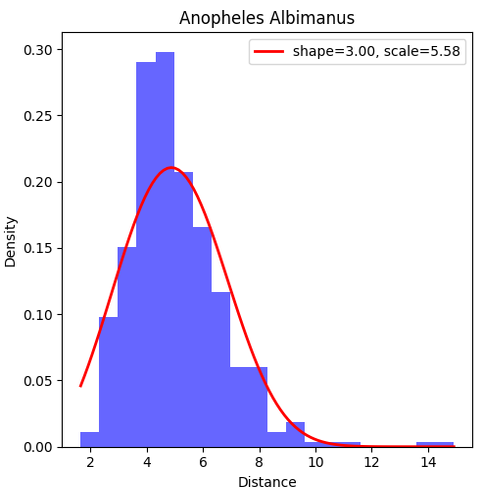}
    \caption{An. Albimanus}
  \end{subfigure}
  \begin{subfigure}{0.19\textwidth}
    \centering
    \includegraphics[width=\linewidth]{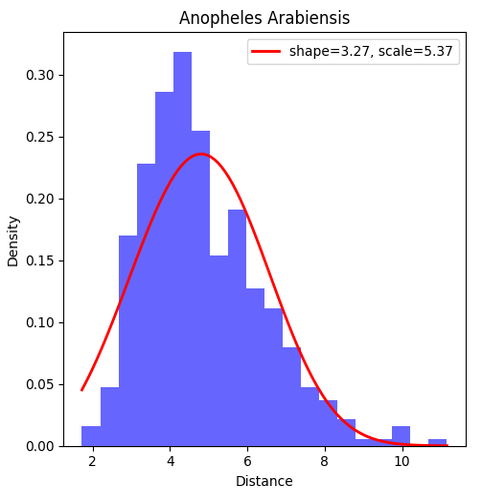}
    \caption{An. Arabiensis }
  \end{subfigure}
  \begin{subfigure}{0.19\textwidth}
    \centering
    \includegraphics[width=\linewidth]{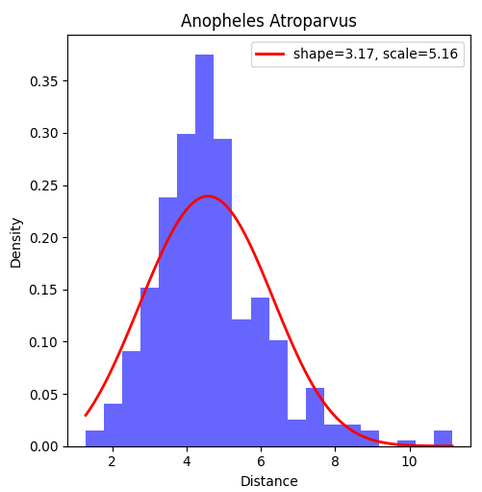}
    \caption{An. Atroparvus}
  \end{subfigure}
  
  \begin{subfigure}{0.19\textwidth}
    \centering
    \includegraphics[width=\linewidth]{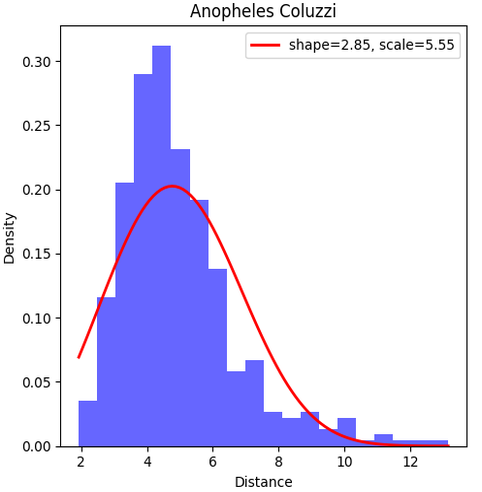}
    \caption{An. Coluzzi}
  \end{subfigure}
  \begin{subfigure}{0.19\textwidth}
    \centering
    \includegraphics[width=\linewidth]{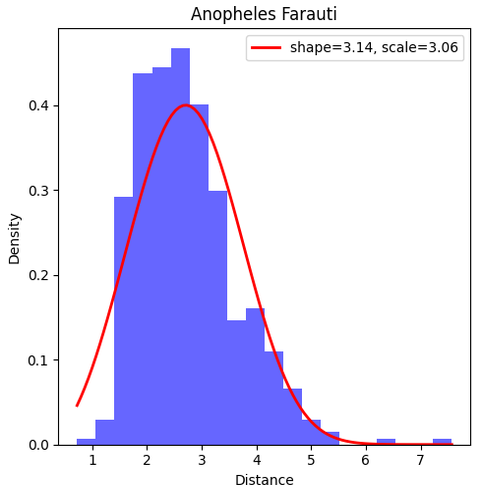}
    \caption{An. Farauti}
  \end{subfigure}
  \begin{subfigure}{0.19\textwidth}
    \centering
    \includegraphics[width=\linewidth]{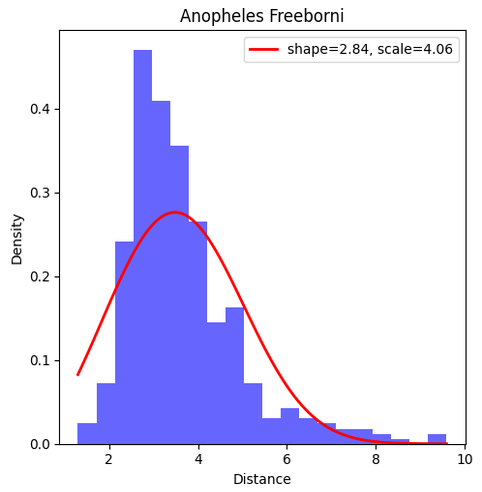}
    \caption{An. Freeborni}
  \end{subfigure}
  \begin{subfigure}{0.19\textwidth}
    \centering
    \includegraphics[width=\linewidth]{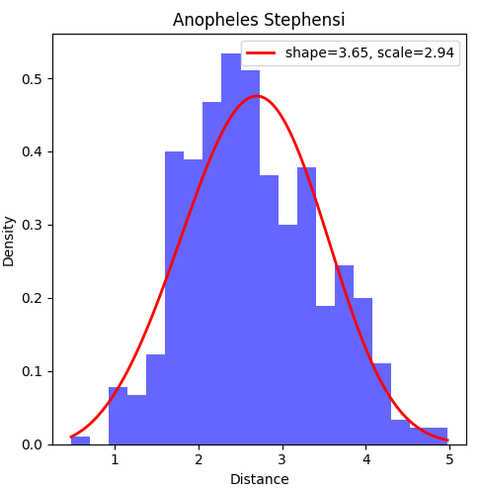}
    \caption{An. Stephensi}
  \end{subfigure}
  \begin{subfigure}{0.19\textwidth}
    \centering
    \includegraphics[width=\linewidth]{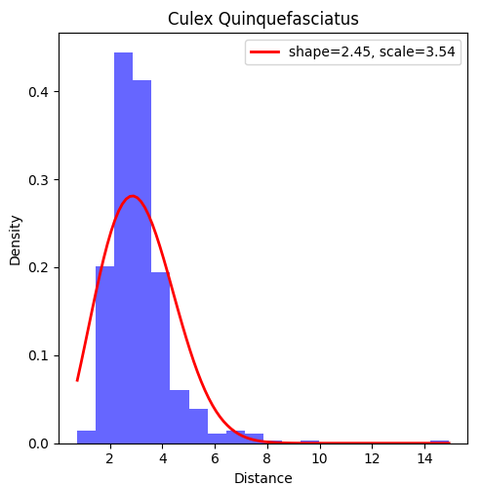}
    \caption{Cx. Quinquefasciatus}
  \end{subfigure}
  \caption{Weibull distribution for Swin-B model containing all mosquito classes on the Extended Test Set images.}
  \label{fig:weibull_swin-B}
\end{figure}

Figure~\ref{fig:weibull_swin-B} displays the Weibull distribution plots, obtained from (\ref{Weibull}), for all ten classes identified by the Swin-S model. Each subplot corresponds to a different mosquito species, demonstrating how the model's predictions follow a Weibull distribution for each class. The Weibull distributions' consistent shape and scale parameters suggest that the Swin-S model effectively captures the underlying data distribution for each class, indicating robust performance in classifying these species.

\begin{figure}[htbp]
  \centering
  
  \begin{subfigure}[b]{0.45\textwidth}
    \includegraphics[width=\textwidth]{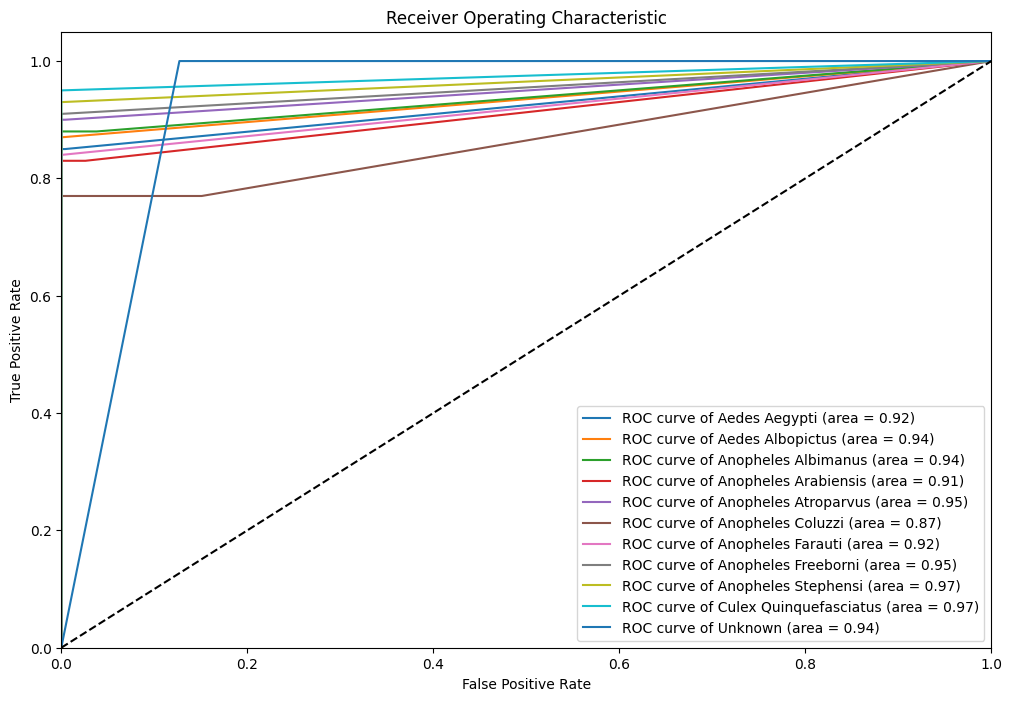}
    \caption{Swin-B}
    \label{swingsubfig2}
  \end{subfigure}
  \hfill
  \begin{subfigure}[b]{0.45\textwidth}
    \includegraphics[width=\textwidth]{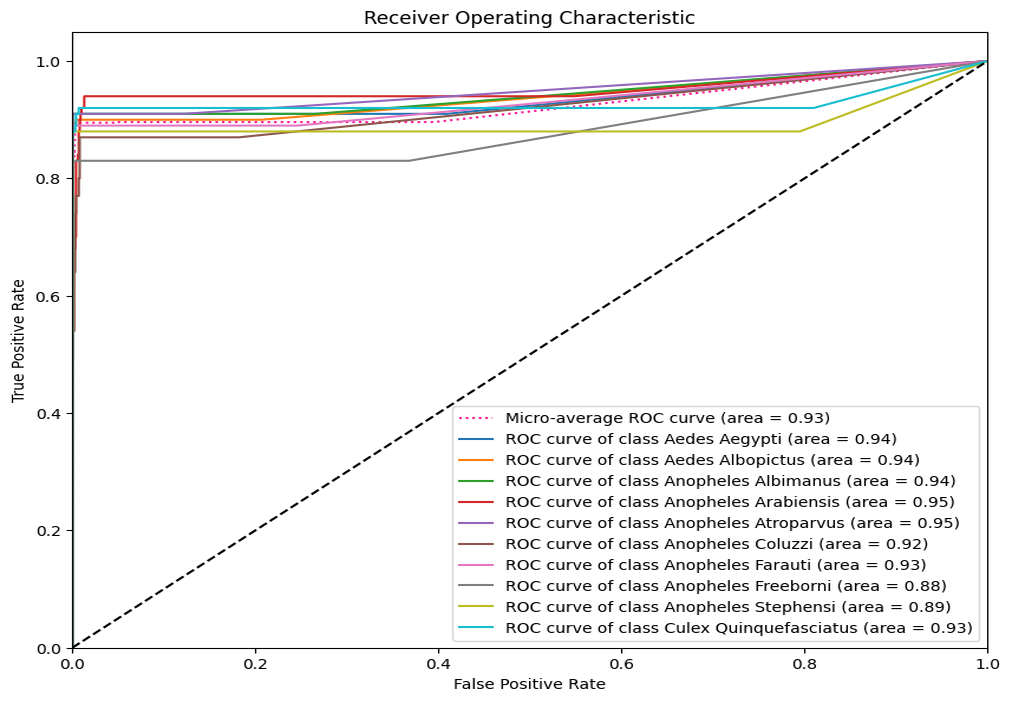}
    \caption{Xception}
    \label{swingsubfig4}
  \end{subfigure}
  \hfill
  \begin{subfigure}[b]{0.45\textwidth}
    \includegraphics[width=\textwidth]{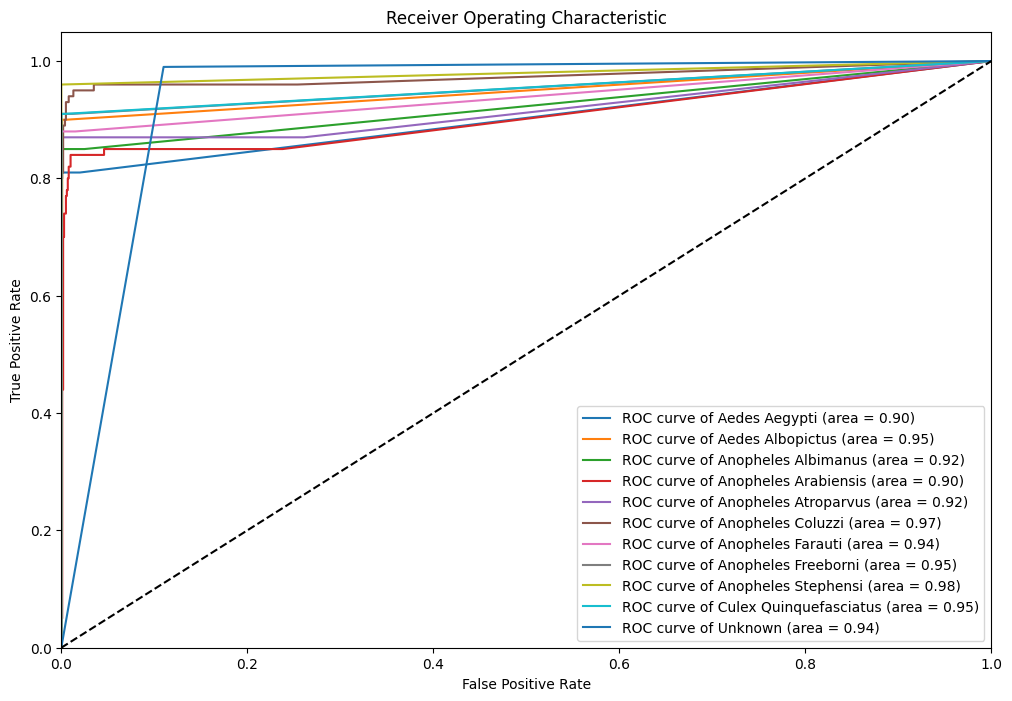}
    \caption{ViT}
    \label{swingsubfig6}
  \end{subfigure}
  \hfill
  \begin{subfigure}[b]{0.45\textwidth}
    \includegraphics[width=\textwidth]{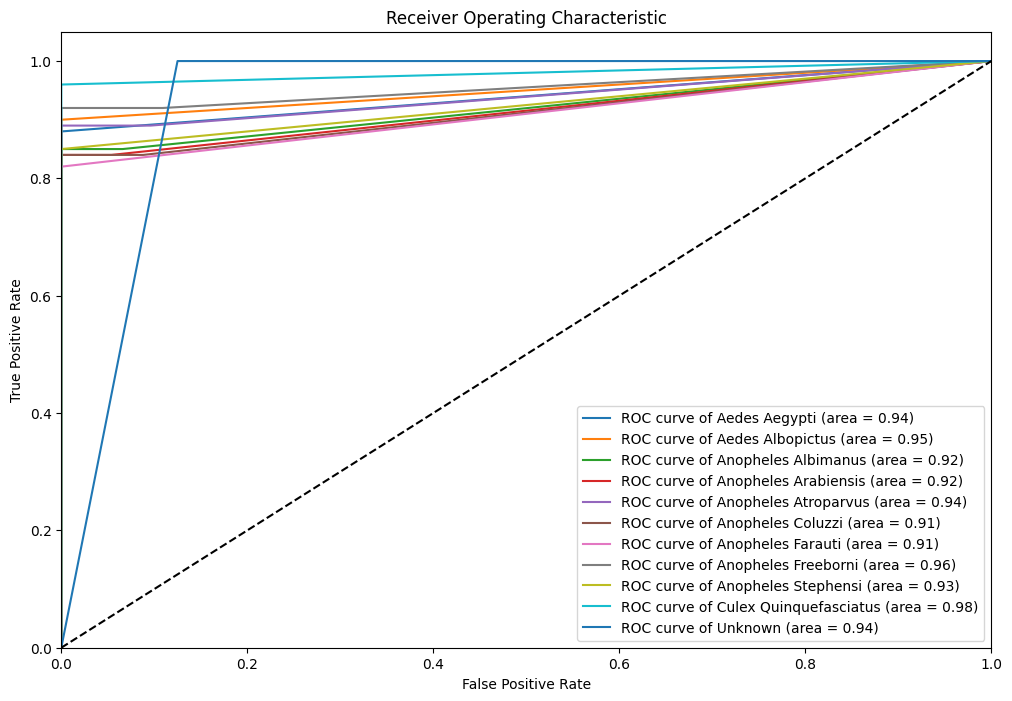}
    \caption{Swin-S}
    \label{swingsubfig8}
  \end{subfigure}
  \caption{ROC curves of different models for open-set learning with OpenMax using the Extended Test Set images.}
  \label{fig:swingraph_1}
\end{figure}

Figure~\ref{fig:swingraph_1} showcases the ROC curves for different models (Swin-B, Xception, ViT, and Swin-S) in the context of open-set learning with OpenMax and Weibull distribution. The visual comparison indicates that OpenMax generally enhances the models' robustness in open-set scenarios by improving their true positive rates while managing false positives. Each class, including the micro-average, achieves an Area Under the Curve (AUC) of 1.00, indicating perfect classification performance. This condition suggests that the model is highly effective at distinguishing between known and unknown classes, demonstrating its robust capability in an open-set recognition scenario.

Table~\ref{threshold_Swin-B} presents the performance metrics of the applied models (Swin-B, Xception, ViT, ResNet50, and Swin-S) across various thresholds, calculated from (\ref{OpenMax}), while implementing the OpenMax technique. Among the models, Xception demonstrates the best performance, achieving the highest accuracy of 90\% and an F1 score of 0.90 at the 0.10 threshold along with the highest macro and weighted average F1 score of 0.91, indicating its superior capability in handling open-set recognition tasks compared to the other models.

\begin{table}[h]
\centering
\caption{Performance metrics of the applied models for various OpenMax thresholds}
\label{threshold_Swin-B}
\begin{adjustbox}{max width=\textwidth}
\begin{tabular}{c|ccc|ccc|ccc}
\hline
\specialrule{2pt}{0pt}{0pt} 
\multirow{2}{*}{Model}  & \multirow{2}{*}{Threshold} & \multirow{2}{*}{Accuracy} & \multirow{2}{*}{F1 Score} & \multicolumn{3}{c|}{Macro Average} & \multicolumn{3}{c}{Weighted Average} \\ \cline{5-10} 
                        &                            &                           &                           & Precision   & Recall  & F1 Score  & Precision    & Recall   & F1 Score   \\ \hline
\multirow{5}{*}{Swin-B} & 0.5                        & 0.54                      & 0.54                      & 0.92        & 0.54    & 0.63      & 0.92         & 0.54     & 0.63       \\ \cline{2-10} 
                        & 0.4                        & 0.65                      & 0.65                      & 0.93        & 0.65    & 0.72      & 0.93         & 0.65     & 0.72       \\ \cline{2-10} 
                        & 0.3                        & 0.73                      & 0.73                      & 0.93        & 0.73    & 0.79      & 0.93         & 0.73     & 0.79       \\ \cline{2-10} 
                        & 0.2                        & 0.81                      & 0.81                      & 0.94        & 0.81    & 0.85      & 0.94         & 0.81     & 0.85       \\ \cline{2-10} 
                        & 0.1                        & 0.88                      & 0.88                      & 0.95        & 0.88    & 0.90      & 0.95         & 0.88     & 0.90       \\ \specialrule{2pt}{0pt}{0pt} 
                        \multirow{5}{*}{Xception} & 0.5                        & 0.60                      & 0.60                      & 0.93        & 0.60    & 0.68      & 0.93         & 0.60     & 0.68       \\ \cline{2-10} 
                        & 0.4                        & 0.68                      & 0.68                      & 0.93        & 0.68    & 0.75      & 0.93         & 0.68     & 0.75       \\ \cline{2-10} 
                        & 0.3                        & 0.77                      & 0.77                      & 0.93        & 0.77    & 0.81      & 0.93         & 0.77     & 0.81       \\ \cline{2-10} 
                        & 0.2                        & 0.84                      & 0.84                      & 0.94        & 0.84    & 0.87      & 0.94         & 0.84     & 0.87       \\ \cline{2-10} 
                        & \textbf{0.1}                        & \textbf{0.90}                      & \textbf{0.90}                      & 0.94        & 0.90    & 0.91      & 0.94         & 0.90     & 0.91       \\ \specialrule{2pt}{0pt}{0pt} 
                        \multirow{5}{*}{ViT} & 0.5                        & 0.53                      & 0.53                      & 0.92        & 0.53    & 0.62      & 0.92         & 0.53     & 0.62       \\ \cline{2-10} 
                        & 0.4                        & 0.64                      & 0.64                      & 0.92        & 0.64    & 0.71      & 0.92         & 0.64     & 0.71       \\ \cline{2-10} 
                        & 0.3                        & 0.74                      & 0.74                      & 0.93        & 0.74    & 0.89      & 0.93         & 0.74     & 0.89       \\ \cline{2-10} 
                        & 0.2                        & 0.82                      & 0.82                      & 0.93        & 0.82    & 0.85      & 0.93         & 0.82     & 0.85       \\ \cline{2-10} 
                        & 0.1                        & 0.89                      & 0.89                      & 0.94        & 0.89    & 0.91      & 0.94         & 0.89     & 0.91       \\ \specialrule{2pt}{0pt}{0pt} 
                        \multirow{5}{*}{ResNet50} & 0.5                        & 0.57                      & 0.57                      & 0.86        & 0.57    & 0.64      & 0.86         & 0.57     & 0.64       \\ \cline{2-10} 
                        & 0.4                        & 0.66                      & 0.66                      & 0.86        & 0.66    & 0.71      & 0.86         & 0.66     & 0.71       \\ \cline{2-10} 
                        & 0.3                        & 0.74                      & 0.74                      & 0.86        & 0.74    & 0.77      & 0.86         & 0.77     & 0.74       \\ \cline{2-10} 
                        & 0.2                        & 0.79                      & 0.79                      & 0.87        & 0.79    & 0.81      & 0.87         & 0.79     & 0.81       \\ \cline{2-10} 
                        & 0.1                        & 0.85                      & 0.85                      & 0.88        & 0.85    & 0.86      & 0.88         & 0.85     & 0.86       \\ \specialrule{2pt}{0pt}{0pt} 
                        \multirow{5}{*}{Swin-S} & 0.5                        & 0.58                      & 0.58                      & 0.93        & 0.58    & 0.66      & 0.93         & 0.58     & 0.66       \\ \cline{2-10} 
                        & 0.4                        & 0.68                      & 0.68                      & 0.93        & 0.68    & 0.75      & 0.93         & 0.68     & 0.75       \\ \cline{2-10} 
                        & 0.3                        & 0.76                      & 0.76                      & 0.93        & 0.76    & 0.81      & 0.93         & 0.76     & 0.81       \\ \cline{2-10} 
                        & 0.2                        & 0.83                      & 0.83                      & 0.94        & 0.83    & 0.86      & 0.94         & 0.83     & 0.86       \\ \cline{2-10} 
                        & 0.1                        & 0.89                      & 0.89                      & 0.95        & 0.89    & 0.90      & 0.95         & 0.89     & 0.90       \\ \specialrule{2pt}{0pt}{0pt} 
\end{tabular}
\end{adjustbox}
\end{table}

\begin{figure}[h]
  \centering
  \includegraphics[width=0.4\linewidth]{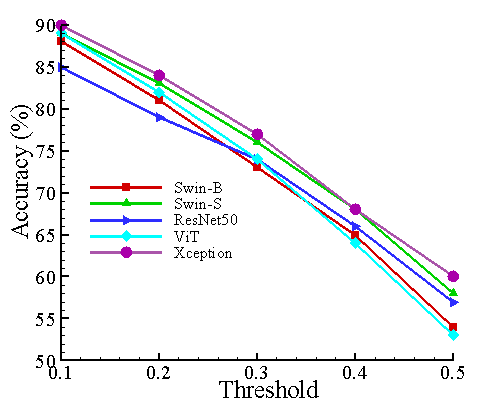}
  \caption{OpenMax threshold vs. accuracy for various models}
  \label{fig:threshold}
\end{figure}

Figure~\ref{fig:threshold} depicts the relationship between OpenMax threshold values and accuracy for various models, including Swin-B, Swin-S, ResNet50, ViT, and Xception. Among these models, the Xception model exhibits the highest accuracy at a threshold value of 0.10, achieving an accuracy of 90\%. As the threshold value increases from 0.10 to 0.50, all models show a general decline in accuracy, with the Xception model consistently performing better at lower thresholds.

\begin{figure}[htbp]
  \centering
  \begin{subfigure}{0.19\textwidth}
    \centering
    \includegraphics[width=\linewidth]{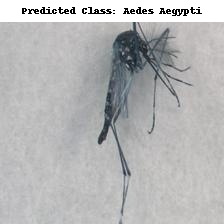}
    \caption{Ae. Aegypti}
  \end{subfigure}
  \begin{subfigure}{0.19\textwidth}
    \centering
    \includegraphics[width=\linewidth]{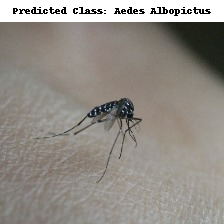}
    \caption{Ae. Albopictus}
  \end{subfigure}
  \begin{subfigure}{0.19\textwidth}
    \centering
    \includegraphics[width=\linewidth]{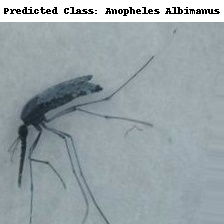}
    \caption{An. Albimanus}
  \end{subfigure}
  \begin{subfigure}{0.19\textwidth}
    \centering
    \includegraphics[width=\linewidth]{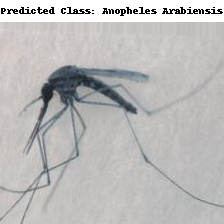}
    \caption{An. Arabiensis}
  \end{subfigure}
  \begin{subfigure}{0.19\textwidth}
    \centering
    \includegraphics[width=\linewidth]{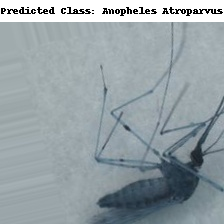}
    \caption{An. Atroparvus}
  \end{subfigure}
  
  \begin{subfigure}{0.19\textwidth}
    \centering
    \includegraphics[width=\linewidth]{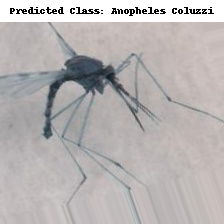}
    \caption{An. Coluzzi}
  \end{subfigure}
  \begin{subfigure}{0.19\textwidth}
    \centering
    \includegraphics[width=\linewidth]{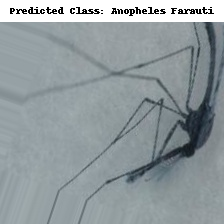}
    \caption{An. Farauti}
  \end{subfigure}
  \begin{subfigure}{0.19\textwidth}
    \centering
    \includegraphics[width=\linewidth]{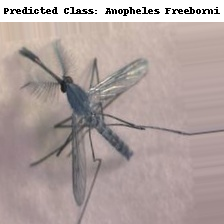}
    \caption{An. Freeborni}
  \end{subfigure}
  \begin{subfigure}{0.19\textwidth}
    \centering
    \includegraphics[width=\linewidth]{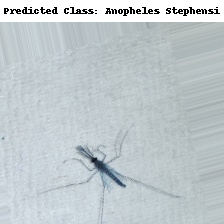}
    \caption{An. Stephensi}
  \end{subfigure}
  \begin{subfigure}{0.19\textwidth}
    \centering
    \includegraphics[width=\linewidth]{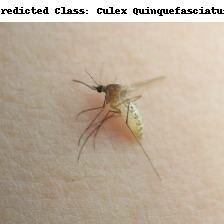}
    \caption{Cx. Quinquefasciatus}
  \end{subfigure}
  \par\bigskip
  \begin{subfigure}{0.19\textwidth}
    \centering
    \includegraphics[width=\linewidth]{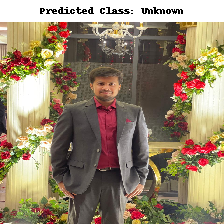}
    \caption{Human}
  \end{subfigure}
  \begin{subfigure}{0.19\textwidth}
    \centering
    \includegraphics[width=\linewidth]{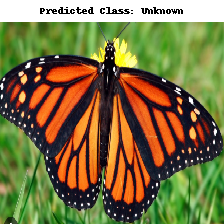}
    \caption{Butterfly}
  \end{subfigure}
  \begin{subfigure}{0.19\textwidth}
    \centering
    \includegraphics[width=\linewidth]{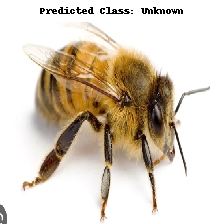}
    \caption{Bee}
  \end{subfigure}
  \begin{subfigure}{0.19\textwidth}
    \centering
    \includegraphics[width=\linewidth]{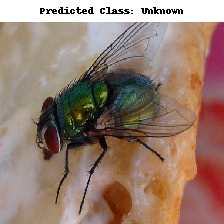}
    \caption{Fly}
  \end{subfigure}
  \begin{subfigure}{0.19\textwidth}
    \centering
    \includegraphics[width=\linewidth]{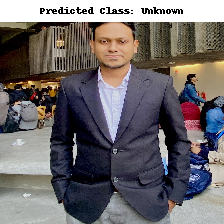}
    \caption{Human}
  \end{subfigure}
  \caption{Classifications of the Xception model for open-set learning with OpenMax.}
  \label{fig:xception-BOM}
\end{figure}

Figure~\ref {fig:xception-BOM} illustrates the classification results of the Xception model, which performed the best among the models enhanced with the OpenMax technique on an open-set dataset comprising 11 classes, including both known (10 mosquito species) and unknown categories. Each subfigure depicts an image from the dataset with its corresponding actual and predicted class labels, highlighting the model's ability to handle open-set recognition by identifying classes not seen during training, such as \enquote{Human}, \enquote{Butterfly}, \enquote{Bee}, and \enquote{Fly} as \enquote{Unknown}. This situation demonstrates the effectiveness of OpenMax in developing the robustness of the Xception model by enabling it to recognize and manage previously unseen classes, thus reducing the likelihood of misclassification in a real-world scenario.

Table~\ref{compareM} compares the proposed mosquito classification system with related studies. The proposed system, which merges four public datasets totaling 5000 images and ten mosquito classes and uses the Swin-B model, achieves the highest accuracy of 99.6\%. 

\begin{table}[h]
\caption{Comparison of the proposed mosquito classification system with related studies}
\label{compareM}
\centering
\begin{adjustbox}{max width=\textwidth}
\begin{tabular}{llllll}
\hline
Study       & Dataset and Number of Images After Augmentation                 & Number of Classes & Best Model & Accuracy (\%) & F1 Score \\ \hline
\cite{rustam2022vector}   & IEEE data port          & 2     & ETC & 99.20\%  & 0.99                        \\ 
\cite{isawasan2023aprotocol}    & Open-source, 2400 images      & 2     & VGG16 & 98.25\% & 0.95                  \\ 
\cite{siddiqui2023transfer}   & Custom, 5400 images   & 6     & VGG16 & 91.17\% & N/A                      \\ 
\cite{zhao2022swin}    & 9900 images             & 7     & Swin Transformer & 96.30\% & 0.98        \\ 
\cite{omucheni2023rapid}   & Custom, 397 images    & 2     & QSVM & 93\% & N/A                          \\ 
\cite{goodwin2021mosquito}    & N/A                     & 16    & Xception & 97.04$\pm$0.87\% & 0.97   \\ 
\cite{pise2023deep}    & 2648 images             & 3     & GoogLeNet & 96\%  & N/A                     \\ 
\textbf{Proposed}   & \begin{tabular}[c]{@{}l@{}} \textbf{Merged 4 public datasets}\\ \textbf{5000 images}\end{tabular} & \textbf{10}    & \textbf{Swin-B Transformer} & \textbf{99.60\%} & \textbf{0.996}                   \\ \hline

\end{tabular}
\end{adjustbox}
\end{table}

\section{Conclusion}

In this research, transformer-based deep learning techniques have been applied for mosquito classification. A comprehensive dataset comprising ten individual mosquito species (two, seven and one types of Aedes, Anopheles and Culex, respectively) has been constructed from public repositories utilized in related articles. Standard preprocessing and augmentation techniques have been implemented to balance the dataset. The Swin transformer model attains the best performance for closed-set learning in static settings, i.e., test/inference samples' classes have been included in the training set. The lightweight MobileVit technique offers almost equal performance with significantly reduced model complexity and parameter number. The robustness of the applied classification models has been enhanced by applying an open-set learning approach. The Xception deep learning model with the optimized OpenMax threshold demonstrates the best performance in open-set learning. A possible extension of this automatic mosquito classification system can be developing a diverse dataset with a wider variety of mosquito species and genera. Additionally, incorporating more sophisticated open-set and en-world learning techniques, e.g., MetaMax, may improve the system's ability to handle unseen classes, thus increasing its practical applicability in real-world scenarios. In the future, mosquito images and wingbeat sounds with different backgrounds and noises can be merged to develop a multimodal dataset. The proposed classification system can be deployed into smartphones or IoT devices for real-time detection, increasing community participation.


\section*{Conflicts of Interest} 
 \noindent No conflicts of interest.

\section*{Data Availability}

\noindent The dataset and code of this work can be found at \\
\href{https://github.com/AkibCoding/Advanced-Vision-Transformers-and-Open-Set-Learning-for-Mosquito-Classification-.git}
{Advanced-Vision-Transformers-and-Open-Set-Learning-for-Mosquito-Classification}\\


\bibliographystyle{plos2015}
\bibliography{LES}

\end{document}